\pdfoutput=1

\documentclass[11pt]{article}

\usepackage{ACL2023}

\usepackage{times}
\usepackage{latexsym}

\usepackage[T1]{fontenc}

\usepackage[utf8,utf8x]{inputenc}

\usepackage{microtype}

\usepackage{inconsolata}

\usepackage{arydshln} 
\usepackage{booktabs}
\usepackage{subcaption}
\usepackage{graphicx}
\usepackage[encapsulated]{CJK}
\usepackage{amssymb}
\usepackage{amsmath}
\usepackage{xspace}
\usepackage{wrapfig}
\usepackage{enumitem}
\usepackage{colortbl}
\usepackage{multirow}

\usepackage{CJKutf8}

\title{How Vocabulary Sharing Facilitates Multilingualism in LLaMA?}

\author{
Fei Yuan\textsuperscript{\rm1}, Shuai Yuan \textsuperscript{\rm2}, Zhiyong Wu\textsuperscript{\rm1},  Lei Li\textsuperscript{\rm3} \\
\textsuperscript{\rm 1} Shanghai Artificial Intelligence Laboratory \\
\textsuperscript{\rm 2} Hong Kong University of Science and Technology \\
\textsuperscript{\rm 3} Carnegie Mellon University \\
  \texttt{\{yuanfei, wuzhiyong\}@pjlab.org.cn},  \texttt{syuanaf@connect.ust.hk}, \texttt{leili@cs.cmu.edu}
}

\begin{document}
\maketitle

\begin{abstract}

Large Language Models (LLMs), often show strong performance on English tasks, while exhibiting limitations on other languages. What is an LLM's multilingual capability when it is trained only on certain languages? The underlying mechanism remains unclear. This study endeavors to examine the multilingual capability of LLMs from the vocabulary sharing perspective by conducting an exhaustive analysis across 101 languages. Through the investigation of the performance gap before and after embedding fine-tuning, we discovered four distinct quadrants. By delving into each quadrant we provide actionable and efficient guidelines for tuning these languages. Extensive experiments reveal that existing LLMs possess multilingual capabilities that surpass our expectations, and we can significantly improve the multilingual performance of LLMs based on these attributes of each quadrant~\footnote{\url{https://github.com/CONE-MT/Vocabulary-Sharing-Facilitates-Multilingualism}.}.

\end{abstract}

\section{Introduction}

Large Language Models~(LLM), such as GPT~\citep{gpt,openai2023gpt4}, PaLM~\citep{palm}, and LLaMA~\citep{llama1,llama2}, are trained on massive amounts of text data. While these models show strong capabilities on English tasks, their performance in other languages is often limited~\cite{zhu2023multilingual,bang2023multitask}. 

Significant research effort has been dedicated to enhancing multilingual capabilities by using methods such as continued training with abundant monolingual data~\citep{cui2023efficient,yang2023bigtrans}, or employing instruction-tuning~\footnote{Instruction tuning is a method used to train large language models to follow specific instructions to solve a task. We provide an example of instruction tuning format in Appendix~\ref{appendix:it}.} techniques~\citep{zhu2023extrapolating,li2023eliciting}. Despite the encouraging results, the underlying mechanism of LLM's multilingual capability remains mysterious. 

\begin{figure}[!t]
    \centering
    \includegraphics[width=0.48\textwidth]{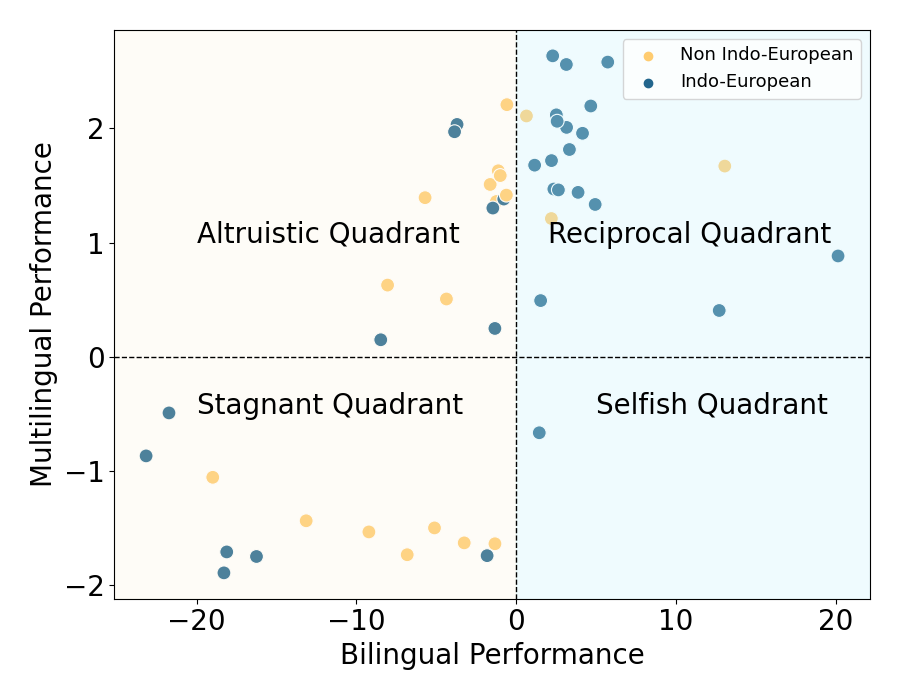}
    \caption{Multilingual capability quadrant. This graph, based on the TED dataset, plots the performance of models fine-tuned with bilingual instructions. Each point represents a model’s performance gain over the original LLaMA. The horizontal axis measures the improvement in bilingual performance, while the vertical axis indicates the enhancement in multilingual performance. }
    \label{fig:quadrant_info}
\end{figure}

Multilingual capability~\cite{lin-etal-2019-choosing} refers to how effectively models that have been fine-tuned in one source language can be applied to tasks in other languages and achieve decent performance. This ability has been extensively studied in machine translation~\citep{johnson2017google, gu2018universal, neubig2018rapid, aharoni2019massively, zhang2020improving} and multilingual pre-trained models~\citep{pires-etal-2019-multilingual,libovický2019languageneutral,wu-dredze-2020-languages}. However, it has not been investigated for English-centric LLMs, given that the pre-training data is predominantly in English. We aim to address this issue by focusing on the multilingual foundation of pre-trained LLMs and providing some guidance to help other people train LLMs more efficiently for non-English languages. Generally, multilingual capabilities are built on two key foundations: the volume of multilingual data used during the pre-training stage~\citep{llama1,llama2,li2023eliciting,Scao2022BLOOMA1}, and the vocabulary~\citep{pires-etal-2019-multilingual,chung2020improving,liang2023xlmv}. In this work, we focus on the latter: vocabulary. 

To investigate the multilingual foundation provided by the vocabulary of an existing LLM, we only fine-tune the embedding layer and keep the rest of the parameters frozen, denoted as Embed FT. This approach requires fewer adjustments to the model parameters than full fine-tuning, and unlike LoRA~\cite{hu2021lora}, it doesn’t require any additional model structure. In our experiments, we focus on the LLaMA as a case study, but the analysis method can be applied to other LLMs.

To examine the multilingual capabilities of LLMs without loss of generality, we applied Embed FT to a 10k en$\rightarrow$x bilingual instruction translation dataset generated by 10k sentences pairs across four distinct datasets: Lego-MT~\cite{legoMT}, Wikimatrix~\citep{schwenk-etal-2021-wikimatrix} and Newscommentary~\citep{tiedemann-2012-parallel}, and Ted~\citep{Ye2018WordEmbeddings}. We evaluated the bilingual performance~(refers to the performance of the fine-tuning languages) and multilingual performance~(refers to the performance of other languages) of each model to determine if there was a significant positive or negative change compared to the original model. From the results, all languages can be categorized into four distinct quadrants. 

The multilingual capability quadrant of the TED dataset, illustrated in Figure~\ref{fig:quadrant_info}, includes four quadrants: the reciprocal quadrant, the altruistic quadrant, the stagnant quadrant, and the selfish quadrant. The full definition of each quadrant is in Section~\ref{sec:inherent_multilingual}. The selfish quadrant refers to scenarios where the fine-tuned model only improves on the fine-tuning language directions but not other languages. It is considered a default quadrant, as languages that fall into the selfish quadrant exhibit behavior that aligns intuitively with the effects of bilingual fine-tuning. 

Certain languages such as Bulgarian fall into the reciprocal quadrant, where training with bilingual data (e.g. English$\rightarrow$Bulgarian) not only enhances bilingual performance but also boosts the multilingual capabilities of other languages. The majority of these languages in this quadrant are from the Indo-European family, benefiting from the pre-training data and vocabulary sharing. For these languages, we find that there is no need to fine-tune all parameters, which could lead to overfitting to a specific language. We recommend fine-tuning only the embedding layer, which yields bilingual performance on par with full fine-tuning while preserving the model’s multilingual capabilities.

Remarkably, certain languages exhibit altruistic characteristics. When we use these languages as training data, their primary effect is to enhance multilingual performance. Upon further analysis, we discovered that the decline in bilingual performance is primarily due to a change in error types: from those that are easy to score to those that are more challenging. The improvement in multilingual performance, on the other hand, stems from vocabulary sharing. For such languages, employing a small dataset for full fine-tuning can be more effective for multilingual capabilities.

Indeed, there are certain languages located in the stagnant quadrant that are quite stubborn. This means that using data from these languages doesn't improve bilingual performance or bring about multilingual benefits. Regardless of parameter-effective tuning strategies (LoRA) or extensive fine-tuning on large datasets, the results are still disappointing. Interestingly, even expanding the vocabulary for full fine-tuning doesn't lead to better results. Then, we find that existing LLMs often over-tokenized these languages, which reduces the density of information they carry. By simply removing the common prefix of tokenized representation, we have seen an average improvement of 2.5 spBLEU points. Our main contributions are:
\begin{itemize}[leftmargin=0.3cm]
    \item We conduct a systematic analysis of the impacts of LLM's vocabulary on their multilingual capabilities, and discover four quadrants based on their embedding fine-tuning performance gap.
    \item We provide practical and efficient technical guides to improve multilingual capabilities for each quadrant.
    \item We perform extensive experiments to verify the effectiveness of quadrant-specific fine-tuning techniques (e.g. 2.5 spBLEU improvement in stagnant quadrant).
\end{itemize}

\section{Background}

\paragraph{Multilingual Large Language Model} Large language models (LLMs;~\citealp{openai2023gpt4,zhang2022opt,gpt,palm,llama1,llama2} have shown demonstrated performance in English, but the performance in other languages is limited. To address this limitation, researchers have proposed multilingual language models (MLLMs) that can handle multiple languages simultaneously. The first line of research proposes to learn a shared representation space for multiple languages by first pre-training on multilingual data and then fine-tuning for specific tasks or languages. Representative works include mBERT~\citep{devlin2019bert}, XLM~\citep{lample2019crosslingual}, XLMR~\citep{conneau2020unsupervised}, BLOOM~\citep{Scao2022BLOOMA1}, XGLM~\citep{lin2022fewshot}, and PolyLM~\citep{wei2023polylm}. Another line of research adopted existing monolingual LLMs to multilingual using techniques such as prompt engineering~\citep{muennighoff2023crosslingual,yong2023prompting}, instruction tuning~\cite{zhu2023extrapolating,li2023eliciting,jiao2023parrot}, or continue training~\citep{cui2023efficient,yang2023bigtrans}. 

\paragraph{The Multilingual Foundation of LLM} The robust multilingual capabilities of LLM are founded on: the presence of diverse multilingual data~\citep{llama1,llama2,li2023eliciting,Scao2022BLOOMA1} and vocabulary~\citep{pires-etal-2019-multilingual,chung2020improving, liang2023xlmv}.

The size of multilingual data is a critical factor in the multilingual capabilities of LLM. LLaMA~\citep{llama1} is pre-trained on a vast scale, with over 1.6 trillion tokens, of which less than 9\% is multilingual data,~\footnote{The original wording (4.5\%) in the LLaMA paper, which only mentioned the inclusion of 20 languages of Wikipedia data. After meticulously checking the datasets involved in the LLaMA pre-training to provide a rigorous account of the quantity of non-English data, we discovered that the Gutenberg dataset includes some multilingual data.} spanning 20 different languages. LLaMA2~\citep{llama2} further enhances the proportion of multilingual data to approximately 11\% and increases the number of languages to around 26. PolyLM~\citep{wei2023polylm} is trained on 640 billion tokens and supports 18 of the most commonly spoken languages. BLOOM~\citep{Scao2022BLOOMA1} is trained with data from 46 natural languages. The existing language data in the pre-training phase provides LLM with a robust foundation for multilingual capabilities.

Another key factor is vocabulary construction. A common approach to constructing vocabulary involves tokenizing text into subwords: including Byte-level Byte-Pair-Encoding (BBPE), Byte-Pair-Encoding (BPE), SentencePiece (SP)~\citep{sennrich-etal-2016-neural,kudo-richardson-2018-sentencepiece,bbpe}, which are units smaller than words that can encapsulate morphological variations. Nevertheless, in a multilingual context encompassing a diverse range of scripts, the base vocabulary comprising subwords can become exceedingly large, leading to inefficiency and sparsity. Further Information on BBPE is in Appendix~\ref{appendix:bbpe}.

\begin{figure*}[!t]
    \centering
    \includegraphics[width=1\textwidth]{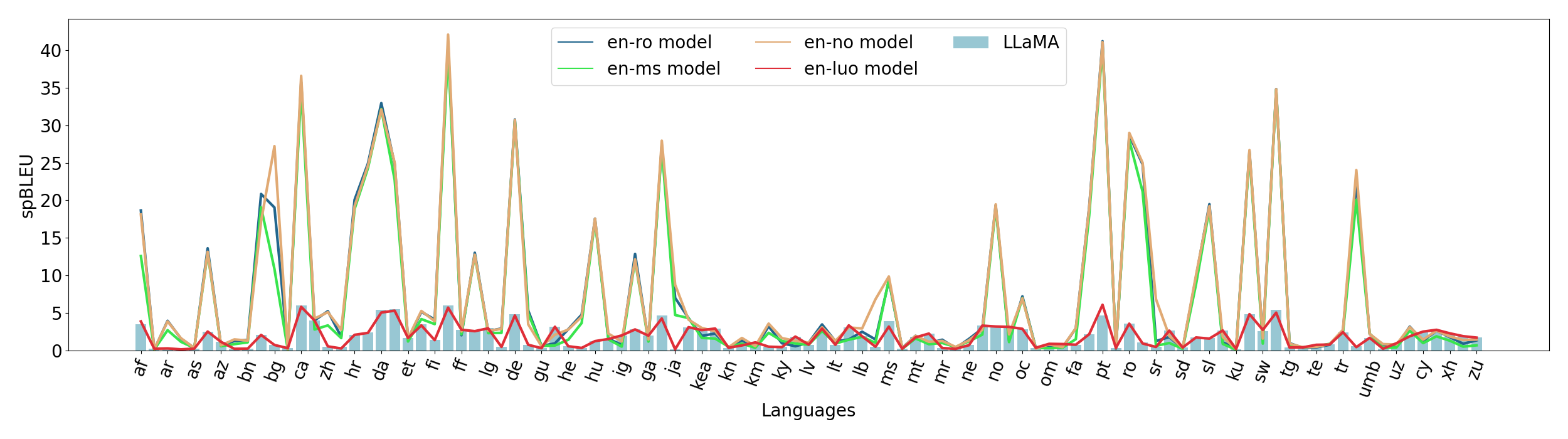}
    \caption{We evaluated the multilingual capabilities of various models on the Flores-101 dataset. The bar graph represents the direct inference results from the original LLaMA, while the line graph illustrates the multilingual performance of models trained on bilingual instruction data from en$\rightarrow$ro, en$\rightarrow$ms, en$\rightarrow$no, and en$\rightarrow$luo. }
    \label{fig:distribution_summary}
\end{figure*}

\section{Inherent Multilingual Capabilities}
\label{sec:inherent_multilingual}

In this section, we begin by exploring the inherent multilingual capabilities of LLMs and give some fascinating observations detailed in Section~\ref{sec:observation}. Drawing on these insights, we then proceed to conduct an in-depth examination of the multilingual capability of LLM in Section~\ref{sec:division}.

\subsection{Observation}
\label{sec:observation}

\paragraph{Setting} We fine-tune a single LLaMA model using en$\rightarrow$x data on the Lego-MT dataset, yielding 101 bilingual-tuned models. 
We train the LLaMA-7B with en$\rightarrow$ro, en$\rightarrow$no, en$\rightarrow$ms, and en$\rightarrow$luo data separately, and then thoroughly evaluate each bilingual-tuned model on all 101 language pairs~(en$\rightarrow$x) to probe its multilingual translation performance on Flores-101's devtest set. Additionally, we follow the same settings for all models throughout the paper. Over 50 bilingual models were tuned using the Wikimatrix and Newscommentary datasets, and more than 55 bilingual models were tuned using the Ted dataset. All these models were trained with identical parameter settings, specifically a learning rate of 2e-5 and a total of 3 epochs, and evaluated bilingual and multilingual performance with beam size = 4.

\paragraph{Phenomena} We observe that LLM demonstrates superior multilingual capabilities far beyond expectation. Some interesting phenomena are:


\paragraph{Phenomenon 1: LLaMA can support additional languages beyond those explicitly mentioned in their pretraining corpus.} In the leftmost part of Figure~\ref{fig:distribution_summary}, it is evident that the bilingual-finetuned en$\rightarrow$ro, en$\rightarrow$ms, and en$\rightarrow$no models exhibit a significant improvement over the original model in en$\rightarrow$af translation. This outcome is quite surprising considering that neither LLaMA's pretraining corpus~\footnote{LLaMA utilizes Wikipedia for its pre-training data which includes 20 languages: bg, ca, cs, da, de, en, es, fr, hr, hu, it, nl, pl, pt, ro, ru, sl, sr, sv, uk.} nor our fine-tuning data contain any text related to af. Similar observations can be made for numerous other languages, as depicted in Figure~\ref{fig:distribution_summary}. This indicates the LLaMA may possess a more robust capability for handling multiple languages than previously expected. 




\paragraph{Phenomenon 2: The performance distribution of bilingual-tuned models across multiple languages exhibits remarkable consistency.} Intuitively, different bilingual models shall have very different multilingual performance distributions. However, from line plots in Figure~\ref{fig:distribution_summary}, we observe that three bilingual models (en$\rightarrow$ro, en$\rightarrow$no, and en$\rightarrow$ms) showcase an exceptional level of consistency. We speculate such a phenomenon might be caused by a similar instruction-tuning process. However, further experiments on the en$\rightarrow$luo model reject the above hypothesis. Therefore, we hypothesize that such a phenomenon only occurs in certain languages and might be related to certain underexplored mechanisms. 

\subsection{Quantify Multilingual Capability at Scale}
\label{sec:division}
Given the phenomena above, we scale our evaluation to more languages to validate our findings. 

\paragraph{Setting} We conduct experiments on LLaMA with 3 epochs on all 101 language pairs(en$\rightarrow$x) in Flores-101 using parallel multilingual corpora. For each language pair, we sample at most 10k sentence pairs at random unless it has fewer than 10k sentence pairs. We then train models using the Embed FT method. For evaluation on Flores-101's devtest including 12 respective languages (details provided in  Appendix~\ref{appendix:appendix_division}), we use a beam size of 4 and spBLEU~(SentencePiece BLEU) as the metric. 

\paragraph{Observation} Inspecting the large-scale evaluation results, we make the following observation: some bilingual models exhibit highly similar yet surprising behaviors. As a counterintuitive example, we showcase a group of ``selfless'' bilingual models (see Table~\ref{tab:more_indo_lgs}). In common belief, fine-tuning LLM on one language pair shall definitely improve its performance. However, to our surprise, fine-tuning these ``selfless'' bilingual models(column LG) might even hurt their performance(comparing en$\rightarrow$LG column with LLaMA column). What's even more interesting is that the multilingual performance of these models is significantly improved.


\begin{table}[!h]
    \centering
    \footnotesize
    \resizebox{0.95\linewidth}{!}{
 \begin{tabular}{c|c|c|cc|c|c}
    \toprule
\textbf{Type} & \textbf{LG}  & \textbf{LLaMA}  & \textbf{en$\rightarrow$af}  & \textbf{en$\rightarrow$ro}  & \textbf{en$\rightarrow$LG}  & \textbf{Multilingual} \\
\midrule
& af  & 3.5  & 15.6  & 20.0  & 15.6  & 17.8 \\
& ro  & 3.6  & 18.6  & 28.7  & 28.7  & 23.7 \\
\midrule
\multirow{12}{*}{selfless} & ln  & 2.9  & 7.9  & 20.9  & 0.9  & 14.4 \\
& ns  & 3.3  & 7.9  & 22.6  & 1.4  & 15.3 \\
& lo  & 1.8  & 8.7  & 17.8  & 0.1  & 13.3 \\
& km  & 1.1  & 9.7  & 21.3  & 0.1  & 15.5 \\
& ig  & 2.0  & 9.7  & 19.8  & 1.2  & 14.7 \\
& ps  & 0.9  & 8.9  & 17.2  & 0.5  & 13.1 \\
& my  & 0.3  & 11.2  & 22.8  & 0.0  & 17.0 \\
& lv  & 0.7  & 10.5  & 22.5  & 0.4  & 16.5 \\
& xh  & 2.3  & 9.4  & 21.7  & 2.0  & 15.5 \\
& mn  & 0.2  & 12.0  & 22.8  & 0.0  & 17.4 \\
& am  & 0.2  & 8.3  & 14.9  & 0.0  & 11.6 \\
& pa  & 0.3  & 8.8  & 18.8  & 0.1  & 13.8 \\
    \bottomrule
    \end{tabular}
    }
    \caption{Consistent performance gains in translation across multiple languages. Each row represents a model that has been trained using en$\rightarrow$LG bilingual dataset. Multilingual performance refers to the average result of en$\rightarrow$af and en$\rightarrow$ro.}
    \label{tab:more_indo_lgs}
\end{table}


To quantitatively investigate the language clustering behavior, as well as dig the root of the phenomena mentioned above, we propose to categorize languages into four quadrants using a two-dimensional Cartesian system. As shown in Figure 1, the x-axis represents bilingual performance, and the y-axis represents multilingual performance. Before clustering, we first establish a categorization criteria.

\paragraph{Criteria} We use the bilingual/multilingual performance changes before and after fine-tuning to measure whether the tuning results in gain or loss:
\begin{equation}
        \Delta_{\mathrm{lg}} = 
    \begin{cases}
        \frac{P_{\text{post}} }{P_{\text{pre}}} - 2,& \text{if } P_{\text{pre}}\geq T\\
         \frac{P_{\text{post}} - 2T}{P_{\text{pre}}},              & \text{otherwise}
    \end{cases}
    \label{eq:quadrant_definition}
\end{equation}
where the $P_{\text{post}}$ represents the translation performance after fine-tuning, $P_{\text{pre}}$ indicates the performance before the fine-tuning process, $T$ serves as a threshold for smoothing, and 2 is a hyperparameter quantifies the extend for significant changes. We select based on a preliminary study, for further information see Appendix~\ref{appendix:hyper_selection}. The calculation of $\Delta_{\mathrm{lg}}$ for bilingual performance is straightforward, for the multilingual performance, we consider the average performance of en$\rightarrow$af and en$\rightarrow$ro translations. This is primarily due to our observation that changes in multilingual performance are significantly mirrored in that of en$\rightarrow$af and en$\rightarrow$ro, details are in Appendix~\ref{appendix:appendix_division}.

\begin{table*}[!t]
    \centering
    \footnotesize
    \resizebox{0.95\textwidth}{!}{
    \begin{tabular}{p{1.5cm}|p{6.5cm}|p{3.5cm}|p{6cm}|p{3.5cm}}
    \toprule
    \textbf{Dataset} & \textbf{ Reciprocal Quadrant} & \textbf{Selfish Quadrant} & \textbf{Altruistic Quadrant} & \textbf{Stagnant Quadrant}\\
    \midrule
    Lego-MT &  af, bs, bg, ca, hr, cs, da, mk, ms, no, oc, pl, pt, ro, sk, sl & ast, be, tl, fr, gl, de, hu, id, it, ky, lt, ml, mt, mi, ny, fa, ru, sr, es, sw, sv, tg, uk & am, ar, hy, as, bn, my, ceb, zh, et, fi, gu, he, is, ig, ga, jv, km, ko, lo, lv, ln, mr, mn, ne, ns, ps, pa, sd, so, tr, ur, uz, vi, cy, xh, zu & te, zhtrad, ff, lg, el, ja, kam, kk, luo, lb, or, om, sn, ku, ta, th, umb, wo, yo \\
    \midrule
    New & bs, bg, ca, hr, cs, da, nl, fr, gl, de, el, hi, hu, id, it, ja, mk, no, pl, pt, ro, ru, sr, sk, sl, es, sv, uk &  & ar, az, be, zh, et, tl, fi, ka, he, is, jv, kk, ko, lt, lb, mr, ne, oc, fa, sw, tg, te, tr, vi & bn, ml, ta\\
    \midrule
    Ted &  bg, hr, cs, da, nl, fr, de, el, hu, id, it, ja, mk, pl, pt, ro, ru, sk, sl, es, sv & hi & ar, et, fi, gl, ka, he, ko, lt, mr, fa, sr, th, tr, uk, vi & hy, az, be, bn, bs, my, zh, kk, ms, mn, ku, ta, ur \\
    \midrule
    Summary & \multicolumn{2}{l|}{bg, id, de, ru, da, mk, hu, it, pl, cs, hr, sl, es, sk, sv, ro, pt, fr} & mr, ko, he, fi, et, vi, tr, ar & - \\
    \bottomrule
    \end{tabular}
    }
    \caption{The distribution of various languages across different quadrants. Various factors such as data influence and tuning strategy can lead to instability in some language quadrants. However, we concentrate on languages that demonstrate consistent stability within these quadrants. In the stagnant quadrant, given that different datasets encompass varying numbers of languages, we also take into account the observations.}
    \label{tab:quadrant}
\end{table*}


\paragraph{Quadrant Details} We calculate the above criteria on four multilingual corpora: Lego-MT~\cite{legoMT}, Wikimatrix~\citep{schwenk-etal-2021-wikimatrix} and Newscommentary~\citep{tiedemann-2012-parallel}, and Ted~\citep{Ye2018WordEmbeddings}, and obtain a consistent language classification results as in Table~\ref{tab:quadrant}. The details of datasets and categorization are in Appendix~\ref{appendix:appendix_division}. We summarize the behavior of four quadrants below~(also shown in Figure 1):

\begin{itemize}[leftmargin=0.2cm]
    \item \underline{\textit{Reciprocal Quadrant}}: Models trained on languages from reciprocal quadrant, demonstrate strong bilingual and multilingual performance at the same time.
    \item \underline{\textit{Altruistic Quadrant}}: Models trained on these languages prioritize enhancing others, with minimal impact on their bilingual performance.
    \item \underline{\textit{Stagnant Quadrant}}: Existing tuning strategies appear to have minimal impact on these languages.
    \item \underline{\textit{Selfish Quadrant}}: The selfish quadrant is the most intuitive one: training in a specific language typically improves the performance of that language and merely affects other languages.
\end{itemize}

Please note that the categorization proposed is merely one possibility derived from certain criteria, and there might exist alternatives that lead to slightly different classification results. Nonetheless, We only focus on the consistent classification, produced by Eq.~\ref{eq:quadrant_definition}, across four distinct datasets for our later analysis. We leave the exploration of a better classification metric as future work.

\section{Enhancing Multilingual Capability}
This section conducts a comprehensive analysis of the properties and training strategies of each quadrant to effectively enhance the multilingual capability of LLMs.

\subsection{Reciprocal Quadrant}

Language within the reciprocal quadrant indicates that using any of these languages as training data invariably improves performance in other languages within the same group. We will delve into this relationship to uncover some intriguing insights.

\paragraph{Interpretation: Reciprocal quadrant consists of linguistically similar languages.} The reciprocal quadrant is predominantly occupied by Indo-European languages. These languages are grouped mainly due to their shared vocabulary and grammatical affixes. Furthermore, the original 20 languages supported by LLaMA are predominantly Indo-European, providing a solid foundation. Consequently, tuning one language within the Indo-European family can effectively enhance the performance of other languages within the same family.

\begin{figure}[!t]
    \centering
    \includegraphics[width=0.95\linewidth]{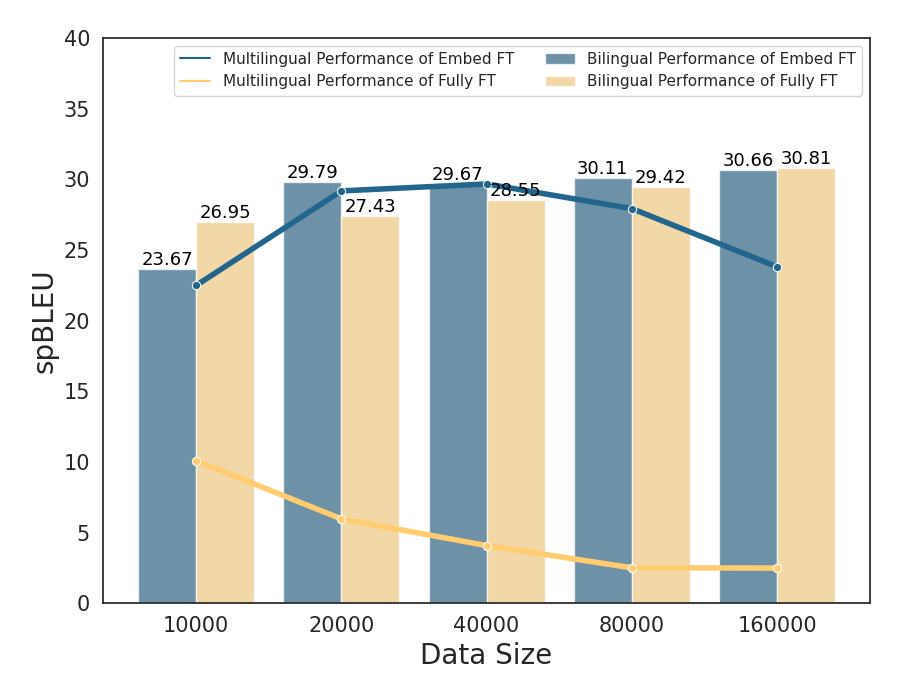}
    \caption{Comparing the Embed FT and Full FT Strategies. In the realm of bilingual performance, both strategies prove equally effective. However, when it comes to multilingual performance, the Embed FT strategy stands out for its adaptability across various languages, while the Full FT strategy tends to over-specialize the model to a single language. The numerical results for each language pair can be found in Appendix~\ref{appendix:single-layer}.}
    \label{fig:first_quadrant}
\end{figure}

\paragraph{Practice Guidance 1: The recommended training strategy for reciprocal languages is Embed FT, which achieves the best performance-generalization trade-off.} Figure~\ref{fig:first_quadrant} illustrates the performance disparity between the models obtained through Embed FT and Full FT strategies under varying amounts of training data. We randomly selected 11 languages from the reciprocal quadrant for testing, including es, pt, ca, de,  da, cs, bg, pl, fr, ru, nl, and averaged the bilingual/multilingual performance across all 11 languages.

For bilingual performance, the Embed FT strategy works as well as the Full FT strategy. As depicted by the bar in Figure~\ref{fig:first_quadrant}, the results indicate that when working with a limited dataset, the model trained by Embed FT demonstrates a slightly inferior performance compared to Full FT. However, as the size of the dataset increases, the model developed using Embed FT not only matches but may even exceed the performance of Full FT.

For multilingual performance, the Embed FT strategy excels in adapting to various languages, while the Full FT strategy tends to make the model overly specialized to a particular language. As illustrated by the line in Figure~\ref{fig:first_quadrant}, the findings suggest that full fine-tuning of a bilingual dataset may lead to overfitting, but this can be effectively mitigated by using the Embed FT strategy.

\begin{table*}[!ht]
    \centering
    \footnotesize
    \resizebox{0.95\linewidth}{!}{
    \begin{tabular}{ccccccccccccccc|c}
        \toprule
        \textbf{\# Lang} & \textbf{Data Size} & \textbf{en$\rightarrow$hr} & \textbf{en$\rightarrow$da} & \textbf{en$\rightarrow$no} & \textbf{en$\rightarrow$ro} & \textbf{en$\rightarrow$ca} & \textbf{en$\rightarrow$cs} & \textbf{en$\rightarrow$bg} & \textbf{en$\rightarrow$pl} & \textbf{en$\rightarrow$es} & \textbf{en$\rightarrow$fr} & \textbf{en$\rightarrow$de} & \textbf{en$\rightarrow$pt} & \textbf{en$\rightarrow$nl} & \textbf{AVG.} \\ 
            \midrule
\rowcolor{lightgray!30}  \multicolumn{16}{c}{\textbf{Bilingual Full Fine-Tuning}} \\
    \midrule
         & 20k & 20.2 & 32.2 & 22.2 & 28.8 & 35.8 & 24.5 & 26.5 & 18.4 & 23.8 & 31.7 & 24.8 & 41.1 & 18.9 & 26.8 \\ 
         & 40k & 21.2 & 32.8 & 24.0 & 29.6 & 37.0 & 25.4 & 27.4 & 18.8 & 25.2 & 34.1 & 25.9 & 41.3 & 22.1 & 28.1 \\ 
         & 80k & 22.4 & 34.8 & 25.6 & 30.8 & 38.5 & 26.4 & 29.3 & 19.1 & 23.6 & 32.9 & 30.8 & 40.6 & 23.5 & 29.1 \\ 
           \midrule
        \rowcolor{lightgray!30}        \multicolumn{16}{c}{\textbf{Multilingual Full Fine-Tuning}} \\
    \midrule
        2 & 160k & 22.9 & 17.2 & 8.7 & 19.0 & 24.9 & 17.8 & 5.1 & 8.7 & 10.7 & 4.5 & 5.4 & 9.6 & 23.7 & 13.7 \\ 
        4 & 40k  & 20.0 & 31.1 & 18.6 & 28.6 & 35.6 & 24.0 & 20.6 & 18.4 & 26.4 & 36.2 & 27.3 & 38.3 & 23.6 & 26.8 \\ 
        8 & 80k & 20.2 & 28.1 & 21.7 & 28.8 & 36.4 & 24.9 & 27.1 & 19.4 & 25.9 & 37.1 & 25.5 & 41.2 & 24.8 & 27.8 \\ 
    \midrule
    \rowcolor{lightgray!30}     \multicolumn{16}{c}{\textbf{Multilingual Embedding Fine-Tuning}} \\
    \midrule
        2 & 160k & 21.5 & 33.1 & 18.5 & 29.5 & 36.0 & 25.6 & 20.5 & 18.8 & 26.9 & 41.8 & 30.7 & 41.5 & 24.8 & 28.4 \\ 
        4 & 40k & 19.9 & 33.3 & 19.2 & 29.7 & 37.1 & 24.9 & 26.7 & 19.6 & 26.8 & 42.8 & 30.8 & 41.0 & 25.3 & 29.0 \\ 
        8 & 80k & 20.3 & 32.8 & 19.2 & 28.6 & 34.6 & 24.6 & 27.0 & 19.1 & 27.0 & 40.0 & 29.9 & 40.7 & 24.5 & 28.3 \\ 
    \bottomrule
    \end{tabular}}
    \caption{Performance comparison of bilingual and multilingual models. In full fine-tuning, multilingual models improve with more languages. However, in embedding fine-tuning, language quantity doesn’t significantly affect performance. Notably, multilingual models slightly underperform compared to bilingual models. In the table, a data size of 80k for 8 languages implies that each language contributes 10k sentence pairs. Out of curiosity about the performance of LLaMA in Indo-European languages, which it does not claim to support, we have added two additional languages, hr and no, during the inference process, based on Guidance 1.}
    \label{tab:more_lgs}
\end{table*}

\paragraph{Practice Guidance 2: While the Full FT model’s multilingual capabilities are influenced by language quantity, the Embed FT model remains unaffected.} Considering Phenomenon 3, which observes a consistent multilingual distribution, we are curious to explore whether a richer language number could bring additional performance gains. To investigate this, we randomly select some languages from the reciprocal quadrant to establish a multilingual setting, and the results of this experiment are displayed in Table~\ref{tab:more_lgs}. In the Full FT, the performance of the multilingual model improves with an increase in the number of languages. However, in the Embed FT, the number of languages does not have a significant impact.

\subsection{Altruistic Quadrant}

Languages that fall into this quadrant demonstrate a "selfless" characteristic. Training based on the data from these languages does not necessarily improve, and may even decrease their performance. Interestingly, it can lead to performance enhancements in other languages. We will conduct a thorough examination of the underlying causes of this phenomenon and propose potential solutions.

\paragraph{Interpretation for bilingual performance decline: The model transitions from an error type that is easy to score to a less score-friendly error type.}  The primary error for LLaMA is ``source copy'', which simply duplicates the source sentence as the translation. This error often leads to moderate scores when there are names, numbers, and punctuation in the translation tasks. However, after tuning, the main error shifts to ``oscillatory hallucination''~\citep{li2023eliciting}, a state where the model becomes stuck in a specific translation state and generates repeated n-grams until it reaches the maximum length. This error makes it challenging to earn the score of spBLEU. Therefore, the performance of the fine-tuned model is lower than that of the original model.

\begin{figure}[!t]
    \centering
    \includegraphics[width=0.95\linewidth]{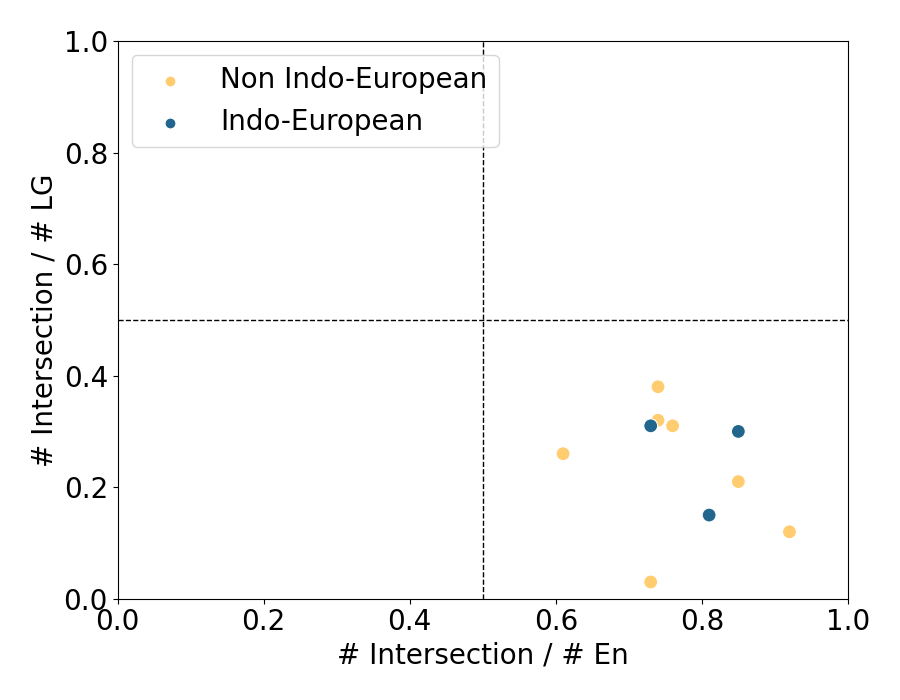}
    \caption{Analyzing linguistics in altruistic languages. A significant overlap in tokenized results with English may enhance performance in Indo-European languages.}
    \label{fig:altruistic_analysis}
\end{figure}

\paragraph{Interpretation for multilingual performance improvement: Those languages’ vocabulary encompasses the majority of English tokens.} We estimate the linguistics of these languages on the Flores-101 benchmark, a multilingual parallel corpus translated by professional translators through a controlled process. For an altruistic language, \textit{LG} we first employ LLaMA's tokenizer to segment the words in both the \textit{LG} and English data from Flores. This allows us to compile the sets of tokens that belong to the \textit{LG} language, denoted as $S_\text{\textit{LG}}$, and the English language, denoted as $S_{\text{\textit{En}}}$. Finally, we calculate the ratios of the size of $S_{\text{\textit{LG}}} \bigcap  S_{\text{\textit{En}}} $ to the size of $S_\text{\textit{LG}}$ and the size of $S_{\text{\textit{En}}}$ respectively. Intriguingly, as shown in Figure~\ref{fig:altruistic_analysis}, we discovered that most tokenized results used in these languages exhibit a high degree of consistency with English.

\paragraph{Practice Guidance: Full FT with a minimal dataset can effectively enhance bilingual performance and maintain a robust multilingual effect.} As shown in Table~\ref{tab:quadrant_strategy_summary}, the altruistic trait is exemplified across different training strategies. However, with Full FT and LoRA, as the dataset size increases, the model tends to overfit the specific language, thereby diminishing its multilingual capabilities. For Embed FT, an increase in data volume does not significantly alter bilingual performance, but it does markedly enhance the multilingual effect. Interestingly, the multilingual effect is not significantly different from that of Full FT with a small dataset. In summary, by employing a small dataset for full fine-tuning, we can strike a balance between bilingual and multilingual performance.

\begin{table}[!t]
    \centering
    \footnotesize    
    \resizebox{0.95\linewidth}{!}{
    \begin{tabular}{cc|cc|cc|cc|cc}

\toprule
    \multirow{2}{*}{\textbf{Setting}} & \multirow{2}{*}{\textbf{Size}} & \multicolumn{2}{c|}{\textbf{en$\rightarrow$vi}} & \multicolumn{2}{c|}{\textbf{en$\rightarrow$tr}} & \multicolumn{2}{c|}{\textbf{en$\rightarrow$ar}} &  \multicolumn{2}{c}{\textbf{AVG.}} \\
    & & B & M & B & M & B & M & B & M \\
\midrule
\multicolumn{2}{c|}{LLaMA} & 1.9 & 3.6 & 2.4 & 3.6 & 0.26 & 3.6 &  1.5 & 3.6 \\
\midrule
\multirow{3}{*}{FT} & 10k & 14.8 & 24.4 & 7.2 & 19.9 & 5.4 & \textbf{24.7} & 9.1	& 23.0\\
& 20k & 18.5 & 22.3 & 8.3 & 9.3 & 6.9 & 22.9  &  11.2 & 18.2\\
& 40k & \textbf{22.3} & 15.9 & \textbf{10.1} & 6.6 & \textbf{9.3} & 21.5 & \textbf{13.9} & 14.7\\
\midrule
\multirow{3}{*}{LoRA} & 10k & 4.9 & \textbf{24.8} & 4.1 & 23.8 & 4.3 & 23.3 & 4.4 & 24.0\\
& 20k & 6.5 & 24.4 & 4.6 & 23.0 & 5.3 & 23.5 & 5.5 & 23.6 \\
& 40k & 7.2 & 18.0 & 5.1 & 17.0 & 5.8 & 21.0 & 6.0 & 18.7\\
\midrule
\multirow{3}{*}{Embed} & 10k & 3.1 & 14.5 & 2.7 & 14.2 & 3.1 & 11.9 & 3.0 & 13.5\\
& 20k & 3.6 & 23.3 & 2.8 & 23.5 & 4.2 & 23.0 & 3.5 & 23.3\\
& 40k & 3.5 & 24.7 & 2.9 & \textbf{24.8} & 4.5 & 23.6 & 3.6 & \textbf{24.4}\\
\bottomrule
    \end{tabular}}
    \caption{The altruistic characteristic is evident in a range of training strategies when trained with the en$\rightarrow$vi, en$\rightarrow$tr, and en$\rightarrow$ar bilingual datasets. Here, “B” denotes the bilingual performance, while “M” signifies the average performance of en$\rightarrow$af and en$\rightarrow$ro.}
    \label{tab:quadrant_strategy_summary}
\end{table}

\begin{table*}[!t]
    \centering
    \footnotesize
    \resizebox{0.87\textwidth}{!}{
    \begin{tabular}{c|c|c|ccccc|ccccc}
    \toprule
    \multirow{2}{*}{\textbf{Setting}} & \multirow{2}{*}{\textbf{Ratio}} & \multirow{2}{*}{\textbf{LLaMA} } & \multicolumn{5}{c|}{\textbf{Full Bilingual Fine-Tuning}} &  \multicolumn{5}{c}{\textbf{LoRA Bilingual Tuning}}  \\
   & & & 10k & 20k & 40K & 80k & 160k & 10k & 20k & 40K & 80k & 160k \\ 
    \midrule
en$\rightarrow$es & \cellcolor{gray!1.7} 1.7 & \cellcolor{blue!4.8} 4.8 & \cellcolor{blue!23.5} 23.5 & \cellcolor{blue!23.8} 23.8 & \cellcolor{blue!25.2} 25.2 & \cellcolor{blue!23.6} 23.6 & \cellcolor{blue!25.9} 25.9 & \cellcolor{blue!26.4} 26.4 & \cellcolor{blue!25.8} 25.8 & \cellcolor{blue!26.6} 26.6 & \cellcolor{blue!26.3} 26.3 & \cellcolor{blue!26.9} 26.9	  \\

en$\rightarrow$pt & \cellcolor{gray!1.9} 1.9 & \cellcolor{blue! 6.0} 6.0 & \cellcolor{blue!41.3} 41.3 & \cellcolor{blue!41.1} 41.1 & \cellcolor{blue!41.3} 41.3 & \cellcolor{blue!40.6} 40.6 & \cellcolor{blue!39.7} 39.7 & \cellcolor{blue!42.0} 42.0 & \cellcolor{blue!42.0} 42.0 & \cellcolor{blue!42.4} 42.4 & \cellcolor{blue!42.0} 42.0 & \cellcolor{blue!41.6} 41.6	  \\

en$\rightarrow$ca & \cellcolor{gray!1.9} 1.9 & \cellcolor{blue!5.7} 5.7 & \cellcolor{blue!34.9} 34.9 & \cellcolor{blue!35.7} 35.7 & \cellcolor{blue!37.0} 37.0 & \cellcolor{blue!38.5} 38.5 & \cellcolor{blue!39.2} 39.2 & \cellcolor{blue!37.3} 37.3 & \cellcolor{blue!37.7} 37.7 & \cellcolor{blue!38.1} 38.1 & \cellcolor{blue!38.6} 38.6 & \cellcolor{blue!39.2} 39.2	  \\

en$\rightarrow$de & \cellcolor{gray!2.0} 2.0 & \cellcolor{blue!4.7} 4.7 & \cellcolor{blue!22.5} 22.5 & \cellcolor{blue!24.8} 24.8 & \cellcolor{blue!25.9} 25.9 & \cellcolor{blue!30.8} 30.8 & \cellcolor{blue!31.2} 31.2 & \cellcolor{blue!27.8} 27.8 & \cellcolor{blue!26.8} 26.8 & \cellcolor{blue!27.3} 27.3 & \cellcolor{blue!31.9} 31.9 & \cellcolor{blue!32.6} 32.6	  \\

en$\rightarrow$no & \cellcolor{gray!2.2} 2.2 & \cellcolor{blue!3.2} 3.2 & \cellcolor{blue!21.2} 21.2 & \cellcolor{blue!22.2} 22.2 & \cellcolor{blue!24.0} 24.0 & \cellcolor{blue!25.6} 25.6 & \cellcolor{blue!28.4} 28.4 & \cellcolor{blue!19.6} 19.6 & \cellcolor{blue!20.1} 20.1 & \cellcolor{blue!21.0} 21.0 & \cellcolor{blue!22.1} 22.1 & \cellcolor{blue!24.0} 24.0	  \\

en$\rightarrow$ro & \cellcolor{gray!2.3} 2.3 & \cellcolor{blue!3.5} 3.5 & \cellcolor{blue!28.3} 28.3 & \cellcolor{blue!28.7} 28.7 & \cellcolor{blue!29.6} 29.6 & \cellcolor{blue!30.8} 30.8 & \cellcolor{blue!34.3} 34.3 & \cellcolor{blue!29.8} 29.8 & \cellcolor{blue!30.0} 30.0 & \cellcolor{blue!30.9} 30.9 & \cellcolor{blue!31.2} 31.2 & \cellcolor{blue!32.7} 32.7	  \\

en$\rightarrow$da & \cellcolor{gray!2.3} 2.3 & \cellcolor{blue!4.9} 4.9 & \cellcolor{blue!31.9} 31.9 & \cellcolor{blue!32.2} 32.2 & \cellcolor{blue!32.8} 32.8 & \cellcolor{blue!34.8} 34.8 & \cellcolor{blue!36.4} 36.4 & \cellcolor{blue!33.4} 33.4 & \cellcolor{blue!34.0} 34.0 & \cellcolor{blue!34.5} 34.5 & \cellcolor{blue!35.3} 35.3 & \cellcolor{blue!36.1} 36.1	  \\

en$\rightarrow$bs & \cellcolor{gray!2.6} 2.6 & \cellcolor{blue!2.0} 2.0 & \cellcolor{blue!23.2} 23.2 & \cellcolor{blue!25.2} 25.2 & \cellcolor{blue!26.5} 26.5 & \cellcolor{blue!28.5} 28.5 & \cellcolor{blue!30.0} 30.0 & \cellcolor{blue!21.7} 21.7 & \cellcolor{blue!22.8} 22.8 & \cellcolor{blue!24.2} 24.2 & \cellcolor{blue!25.0} 25.0 & \cellcolor{blue!25.2} 25.2	  \\

\midrule
en$\rightarrow$gu & \cellcolor{gray!15.0} 15.0 & \cellcolor{blue!0.3} 0.3 & \cellcolor{blue!2.3} 2.3 & \cellcolor{blue!2.2} 2.2 & \cellcolor{blue!4.4} 4.4 & \cellcolor{blue!10.0} 10.0 & \cellcolor{blue!13.2} 13.2 & \cellcolor{blue!1.0} 1.0 & \cellcolor{blue!1.1} 1.1 & \cellcolor{blue!1.5} 1.5 & \cellcolor{blue!1.9} 1.9 & \cellcolor{blue!3.1} 3.1	  \\

en$\rightarrow$kn & \cellcolor{gray!16.9} 16.9 & \cellcolor{blue!0.3} 0.3 & \cellcolor{blue!1.0} 1.0 & \cellcolor{blue!1.5} 1.5 & \cellcolor{blue!3.0} 3.0 & \cellcolor{blue!5.6} 5.6 & \cellcolor{blue!9.9} 9.9 & 
\cellcolor{blue!0.5} 0.5 & \cellcolor{blue!0.4} 0.4 & \cellcolor{blue!0.5} 0.5 & \cellcolor{blue!0.8} 0.8 & \cellcolor{blue!1.0} 1.0	  \\

en$\rightarrow$te & \cellcolor{gray!17.4} 17.4 & \cellcolor{blue!0.7} 0.7 & \cellcolor{blue!4.2} 4.2 & \cellcolor{blue!8.2} 8.2 & \cellcolor{blue!12.8} 12.8 & \cellcolor{blue!17.3} 17.3 & \cellcolor{blue!20.3} 20.3 & \cellcolor{blue!0.6} 0.6 & \cellcolor{blue!0.8} 0.8 & \cellcolor{blue!1.7} 1.7 & \cellcolor{blue!2.9} 2.9 & \cellcolor{blue!5.3} 5.3	  \\

en$\rightarrow$ku & \cellcolor{gray!17.6} 17.6 & \cellcolor{blue!0.2} 0.2 & \cellcolor{blue!0.1} 0.1 & \cellcolor{blue!0.1} 0.1 & \cellcolor{blue!0.2} 0.2 & \cellcolor{blue!0.2} 0.2 & \cellcolor{blue!0.2} 0.2 & \cellcolor{blue!0.2} 0.2 & \cellcolor{blue!0.2} 0.2 & \cellcolor{blue!0.2} 0.2 & \cellcolor{blue!0.2} 0.2 & \cellcolor{blue!0.2} 0.2 \\

en$\rightarrow$my & \cellcolor{gray!21.7} 21.7 & \cellcolor{blue!0.3} 0.3 & \cellcolor{blue!1.0} 1.0 & \cellcolor{blue!2.0} 2.0 & \cellcolor{blue!4.1} 4.1 & \cellcolor{blue!7.3} 7.3 & \cellcolor{blue!9.4} 9.4 & \cellcolor{blue!0.1} 0.1 & \cellcolor{blue!0.1} 0.1 & \cellcolor{blue!0.3} 0.3 & \cellcolor{blue!0.3} 0.3 & \cellcolor{blue!0.4} 0.4	  \\

en$\rightarrow$mr & \cellcolor{gray!38.8} 38.8 & \cellcolor{blue!0.3} 0.3 & \cellcolor{blue!5.0} 5.0 & \cellcolor{blue!7.2} 7.2 & \cellcolor{blue!10.7} 10.7 & \cellcolor{blue!13.7} 13.7 & \cellcolor{blue!15.8} 15.8 & \cellcolor{blue!1.8} 1.8 & \cellcolor{blue!1.9} 1.9 & \cellcolor{blue!2.2} 2.2 & \cellcolor{blue!2.0} 2.0 & \cellcolor{blue!1.7} 1.7 \\

en$\rightarrow$lo & \cellcolor{gray!39.8} 39.8 & \cellcolor{blue!1.8} 1.8 & \cellcolor{blue!1.5} 1.5 & \cellcolor{blue!2.3} 2.3 & \cellcolor{blue!3.7} 3.7 & \cellcolor{blue!7.1} 7.1 & \cellcolor{blue!9.8} 9.8 & \cellcolor{blue!0.4} 0.4 & \cellcolor{blue!0.7} 0.7 & \cellcolor{blue!0.7} 0.7 & \cellcolor{blue!0.6} 0.6 & \cellcolor{blue!0.7} 0.7	  \\

en$\rightarrow$km & \cellcolor{gray!43.0} 43.0 & \cellcolor{blue!1.1} 1.1 & \cellcolor{blue!1.6} 1.6 & \cellcolor{blue!3.1} 3.1 & \cellcolor{blue!6.2} 6.2 & \cellcolor{blue!10.1} 10.1 & \cellcolor{blue!13.4} 13.4 & \cellcolor{blue!0.2} 0.2 & \cellcolor{blue!0.2} 0.2 & \cellcolor{blue!0.5} 0.5 & \cellcolor{blue!0.9} 0.9 & \cellcolor{blue!1.5} 1.5	  \\

    \bottomrule
    \end{tabular}
}
    \caption{The relationship between stagnant languages and the characteristic of over-tokenization. The ``Ratio'' is defined as the number of tokens in a sequence after applying the tokenizer, divided by the sentence length, which is measured by the number of words for space-separated languages and characters. }
    \label{tab:value_range}
\end{table*}

\subsection{Stagnant Quadrant}

Languages in this quadrant exhibit remarkable inertia, as training with their data neither enhances their own performance nor influences the performance of other languages. In this section, we will delve deeper into the inertia phenomenon, examining its potential causes and proposing possible solutions. 

\paragraph{Interpretation: Most languages in the stagnant quadrant are characterized by over-tokenization.} The LLaMA tokenizer, based on the BBPE algorithm, is fundamental for multilingual language processing tasks. Its universal applicability to all languages and the lack of a need for an ‘unknown’ token make it optimal for vocabulary sharing and increase its robustness. Despite being suitable for multilingual learning, BBPE results in byte sequence representation of text that is often much longer (up to 4x) than a character sequence representation. Upon investigation, we find that the over-tokenization phenomenon is prevalent in LLaMA. In an extreme case, a sentence in lo that contains 6 words expands to 352 tokens after tokenization. Additional details in Appendix~\ref{appendix:over-tokenization}.

A comparison between active and stagnant languages, as shown in Table~\ref{tab:value_range}, reveals that:
\begin{itemize}[leftmargin=0.2cm]
    \item activating a stagnant language with full fine-tuning requires more data;
    \item the performance improvement with increasing data is modest; 
    \item certain parameter efficiency fine-tuning strategies, like LoRA, do not affect them.
\end{itemize}

\paragraph{Practice Guidance 1: Expanding the vocabulary is not an effective strategy for stagnant languages. Over-tokenization leads to an increased demand for data.}  When a language is not adequately represented by its vocabulary, the common approach is to expand the lexicon~\citep{tai-etal-2020-exbert,cui2023efficient,ji2023better}. Regrettably, in most instances, this strategy of vocabulary enlargement proves ineffective for stagnant languages. As shown in Table~\ref{tab:extend_vocab}, we present three distinct methods to expand the vocabulary: 

\begin{itemize}[leftmargin=0.2cm]
    \item BBPE~\citep{bbpe}: This follows the approach used in LLaMA for vocabulary construction and involves learning a vocabulary for stagnant language;
    \item BPE~\citep{sennrich-etal-2016-neural}): This utilizes the BPE algorithm and is based on subword units to learn a vocabulary;
    \item SP~\citep{kudo-richardson-2018-sentencepiece}: This learns a vocabulary using the SentencePiece algorithm.
\end{itemize}

Meanwhile, to mitigate potential issues from data quality, we have utilized both MC4 and Flores-101 dev to construct vocabulary.

After training LLaMA on Lego-MT 80k bilingual data, the experimental results indicate that: 
\begin{itemize} [leftmargin=0.2cm]
    \item When there is a substantial amount of data, the impact of data quality on vocabulary expansion can be disregarded;
    \item Among all the vocabulary expansion methods, SP tends to yield better results compared to other solutions;
    \item Almost all vocabulary expansion techniques fail to enhance the performance of stagnant languages significantly.
\end{itemize}

\begin{table}[!t]
    \centering
    \footnotesize    
    \resizebox{1\linewidth}{!}{
    \begin{tabular}{c|cccc|c|cccc}
    \toprule
    \textbf{Source} & \textbf{Type} & \textbf{3k} & \textbf{6k} & \textbf{12k} &  \textbf{Source} & \textbf{Type} & \textbf{3k} & \textbf{6k} & \textbf{12k} \\ 
    \midrule
     \rowcolor{lightgray!30}  \multicolumn{5}{c|}{km - 10.1} & \multicolumn{5}{c}{lo - 7.1} \\
    \midrule
          \multirow{3}{*}{MC4} & BBPE & 5.2 & 3.7 & 2.3  &  \multirow{3}{*}{MC4} & BBPE & 6.2 & 1.7 & 3.6 \\ 
         & BPE & 4.7 & 11.0 & 2.1 &  & BPE & 6.7 & 1.8 & 3.6 \\ 
         & SP & 6.2 & \textbf{11.8} & 10.3 &  & SP & \textbf{7.0} & 6.4 & 4.9 \\ 
         \cdashline{1-10}[0.4pt/2pt]
         \multirow{3}{*}{Flores} & BBPE & 4.6 & 3.5 & 8.5 &  \multirow{3}{*}{Flores} & BBPE & 4.6 & 3.9 & 1.5 \\ 
          & BPE & 4.4 & 3.7 & 8.8 & & BPE & 4.3 & 1.5 & 1.6 \\ 
         & SP & 5.5 & 4.4 & - & & SP & 2.4 & 4.2 & - \\ 
    \midrule
    \rowcolor{lightgray!30}  \multicolumn{5}{c|}{gu - 10.0} & \multicolumn{5}{c}{te - 17.3} \\
    \midrule
         \multirow{3}{*}{MC4} & BBPE & 0.4 & 0.3 & 0.3 &  \multirow{3}{*}{MC4} & BBPE & 9.6 & 8.4 & 6.0 \\ 
         & BPE & 0.4 & 0.2 & 0.3 &    & BPE & 9.7 & 7.7 & 6.7 \\ 
         & SP & 0.3 & 0.2 & 0.4 &  &  SP & \textbf{10.0} & 9.7 & 8.1 \\ 
         \cdashline{1-10}[0.4pt/2pt]
         \multirow{3}{*}{Flores} & BBPE & 0.3 & 0.3 & 0.3 & \multirow{3}{*}{Flores} & BBPE & 9.0 & 8.8 & 7.1 \\ 
         &   BPE & 0.3 & 0.3 & 0.3 & & BPE & 8.9 & 8.2 & 7.2 \\ 
         & SP & \textbf{0.4} & - & - &  & SP & 9.8 & - & - \\ 
    \bottomrule
    \end{tabular}}
    \caption{Exploring various strategies for vocabulary expansion: The term ``km - 10.1'' denotes the bilingual performance (10.1) of full fine-tuning on Lego-MT 80k bilingual data (en$\rightarrow$km) without any vocabulary extension. ``3k'', ``6k'', and ``12k'' refer to the extended vocabulary size. Most vocabulary expansion methods do not significantly enhance the performance of stagnant languages. Due to the limited data in Flores dev, some settings are missing in the table.}
    \label{tab:extend_vocab}
    \vspace{-0.4cm}
\end{table}

\paragraph{Practice Guidance 2: Shortening the subword sequences can significantly boost the performance of stagnant languages.} Given the existence of the over-tokenization problem, we find that among these over-tokenized languages, there are a large amount characters. For example, a Chinese character ``\begin{CJK}{UTF8}{gbsn}{饕}\end{CJK}'' is encoded into three code units ``[227, 234, 260]''. We refer to such characters as ‘over-tokenized characters’ for the sake of simplicity. We then gather all these over-tokenized characters along with their three-byte representations. Interestingly, these over-tokenized characters constitute a significant proportion, about 63.8\%, of the corpus, as indicated in Table~\ref{tab:new_strategy}. Moreover, in the case of over-tokenized languages, all over-tokenized characters begin with the same token~(e.g., 227). Therefore, the obtained three-byte representations are very sparse and result in low information density in representation.

Furthermore, we propose a post-tokenization technique to address the over-tokenization problem. We simply remove the shared prefix of over-tokenization characters and obtain the shortened yet lossless new representations. As a concrete example, we remove \begin{CJK}{UTF8}{gbsn}{饕}\end{CJK}'s prefix [227] from its three-byte representation [227, 234, 260] to get a more compact two-byte representation [234, 260]. Subsequently, we utilized this adjusted representation to train LLaMA on the 80k Lego-MT bilingual dataset. Remarkably, our method outperforms both direct fine-tuning of LLaMA and vocabulary extension, achieving a substantial performance boost with an average of 2.5 points.

\begin{table}[!t]
    \centering
    \footnotesize    
    \resizebox{0.9\linewidth}{!}{
    \begin{tabular}{c|cccc|c}
    \toprule
        \textbf{Setting} & \textbf{en$\rightarrow$km} & \textbf{en$\rightarrow$lo} & \textbf{en$\rightarrow$gu} & \textbf{en$\rightarrow$te} & \textbf{AVG.} \\ 
    \midrule
        Ratio & 47.6\% & 67.0\% & 66.8\% & 73.8\% & 63.8\%\\ 
    \midrule
        Full FT & 10.1 & 7.1 & 10.0 & 17.3 & 11.1 \\
        Extend (Best) & 11.8 & 7.0 & 0.4 & 10.0 & 7.3\\ 
        Our Strategy & \textbf{12.6} & \textbf{9.2} & \textbf{11.3} & \textbf{21.5} & \textbf{13.7}\\
    \midrule
    $\Delta$ & + 2.5 & + 2.1 & + 1.3 & + 4.2 & + 2.6 \\
    \bottomrule
    \end{tabular}}
    \caption{Over-tokenization leads to a decrease in information density for LLM. However, by simply removing the over-tokenized character that shares the same prefix, we can enhance performance, achieving results that surpass both full fine-tuning and vocabulary extension.}
    \label{tab:new_strategy}
\end{table}

\subsection{Guidance Summary}

\paragraph{Guidance for reciprocal languages} For languages, primarily Indo-European languages, situated in the reciprocal quadrant, the optimal strategy is to solely fine-tune the embedding layer and keep the remaining parameters frozen. This is primarily due to these languages having shared vocabulary and grammar rules.

\paragraph{Guidance for altruistic languages} For languages residing in the altruistic quadrant, applying full fine-tuning with a minimal dataset can effectively enhance bilingual performance while maintaining a robust multilingual effect. This is mainly because the tokens in the vocabulary of these languages highly overlap with English. 

\paragraph{Guidance for stagnant languages} Shortening subword sequences can markedly enhance the performance of stagnant languages. Most languages in the stagnant quadrant are over-tokenized, which refers to a situation where a text in this language is typically segmented by a tokenizer into an excessively large number of tokens on average. Expanding the vocabulary does not necessarily enhance the performance of these languages. However, in this paper, we demonstrate that simply removing the identical prefix from over-tokenized characters can significantly improve performance.

We provide more analysis about stagnant languages~(in Appendix~\ref{appendix:over-tokenization}), tuning analysis~(in Appendix~\ref{appendix:single-layer}), and different NLP tasks~(in Appendix~\ref{appendix:analysis}) of LLaMA experiments in the Appendix.

\section{Conclusion} In this study, we performed a comprehensive analysis of 101 languages, categorizing them based on shared characteristics into four distinct quadrants: reciprocal, altruistic, selfish, and stagnant quadrants. Upon examining each quadrant in-depth, we identified the primary reasons for the placement of languages within their respective quadrants and provided some practical guidance for training. However, the primary focus of this study is the analysis of persistent language characteristics within each quadrant. A thorough investigation into the conditions that trigger language migration across various phenomena is a subject for our future research.

\section*{Limitation}

In this paper, we find some interesting phenomena in LLaMA. After expanding our evaluation to include more languages, we found that many of them demonstrated remarkably similar behaviors on translation tasks. Then we grouped them with categorization criteria. Although language classification is not our primary focus, our main interest lies in understanding the reasons behind these classifications and enhancing the multilingual capabilities of LLM. Meanwhile, to delve deeper into the role of Embed FT, we provide a more detailed analysis in Appendix~\ref{appendix:analysis}.

\bibliography{custom}

\begin{thebibliography}{50}
\expandafter\ifx\csname natexlab\endcsname\relax\def\natexlab#1{#1}\fi

\bibitem[{Aharoni et~al.(2019)Aharoni, Johnson, and
  Firat}]{aharoni2019massively}
Roee Aharoni, Melvin Johnson, and Orhan Firat. 2019.
\newblock Massively multilingual neural machine translation.
\newblock \emph{arXiv preprint arXiv:1903.00089}.

\bibitem[{Bang et~al.(2023)Bang, Cahyawijaya, Lee, Dai, Su, Wilie, Lovenia, Ji,
  Yu, Chung, Do, Xu, and Fung}]{bang2023multitask}
Yejin Bang, Samuel Cahyawijaya, Nayeon Lee, Wenliang Dai, Dan Su, Bryan Wilie,
  Holy Lovenia, Ziwei Ji, Tiezheng Yu, Willy Chung, Quyet~V. Do, Yan Xu, and
  Pascale Fung. 2023.
\newblock \href {http://arxiv.org/abs/2302.04023} {A multitask, multilingual,
  multimodal evaluation of chatgpt on reasoning, hallucination, and
  interactivity}.

\bibitem[{Brown et~al.(2020)Brown, Mann, Ryder, Subbiah, Kaplan, Dhariwal,
  Neelakantan, Shyam, Sastry, Askell et~al.}]{gpt}
Tom Brown, Benjamin Mann, Nick Ryder, Melanie Subbiah, Jared~D Kaplan, Prafulla
  Dhariwal, Arvind Neelakantan, Pranav Shyam, Girish Sastry, Amanda Askell,
  et~al. 2020.
\newblock Language models are few-shot learners.
\newblock \emph{Advances in neural information processing systems},
  33:1877--1901.

\bibitem[{Chowdhery et~al.(2022)Chowdhery, Narang, Devlin, Bosma, Mishra,
  Roberts, Barham, Chung, Sutton, Gehrmann et~al.}]{palm}
Aakanksha Chowdhery, Sharan Narang, Jacob Devlin, Maarten Bosma, Gaurav Mishra,
  Adam Roberts, Paul Barham, Hyung~Won Chung, Charles Sutton, Sebastian
  Gehrmann, et~al. 2022.
\newblock Palm: Scaling language modeling with pathways.
\newblock \emph{arXiv preprint arXiv:2204.02311}.

\bibitem[{Chung et~al.(2020)Chung, Garrette, Tan, and
  Riesa}]{chung2020improving}
Hyung~Won Chung, Dan Garrette, Kiat~Chuan Tan, and Jason Riesa. 2020.
\newblock Improving multilingual models with language-clustered vocabularies.
\newblock \emph{arXiv preprint arXiv:2010.12777}.

\bibitem[{Conneau et~al.(2020)Conneau, Khandelwal, Goyal, Chaudhary, Wenzek,
  Guzmán, Grave, Ott, Zettlemoyer, and Stoyanov}]{conneau2020unsupervised}
Alexis Conneau, Kartikay Khandelwal, Naman Goyal, Vishrav Chaudhary, Guillaume
  Wenzek, Francisco Guzmán, Edouard Grave, Myle Ott, Luke Zettlemoyer, and
  Veselin Stoyanov. 2020.
\newblock \href {http://arxiv.org/abs/1911.02116} {Unsupervised cross-lingual
  representation learning at scale}.

\bibitem[{Conneau et~al.(2018)Conneau, Rinott, Lample, Williams, Bowman,
  Schwenk, and Stoyanov}]{xnli}
Alexis Conneau, Ruty Rinott, Guillaume Lample, Adina Williams, Samuel~R.
  Bowman, Holger Schwenk, and Veselin Stoyanov. 2018.
\newblock Xnli: Evaluating cross-lingual sentence representations.
\newblock In \emph{Proceedings of the 2018 Conference on Empirical Methods in
  Natural Language Processing}. Association for Computational Linguistics.

\bibitem[{Cui et~al.(2023)Cui, Yang, and Yao}]{cui2023efficient}
Yiming Cui, Ziqing Yang, and Xin Yao. 2023.
\newblock Efficient and effective text encoding for chinese llama and alpaca.
\newblock \emph{arXiv preprint arXiv:2304.08177}.

\bibitem[{Devlin et~al.(2019)Devlin, Chang, Lee, and
  Toutanova}]{devlin2019bert}
Jacob Devlin, Ming-Wei Chang, Kenton Lee, and Kristina Toutanova. 2019.
\newblock \href {http://arxiv.org/abs/1810.04805} {Bert: Pre-training of deep
  bidirectional transformers for language understanding}.

\bibitem[{Goyal et~al.(2022)Goyal, Gao, Chaudhary, Chen, Wenzek, Ju, Krishnan,
  Ranzato, Guzm{\'a}n, and Fan}]{goyal-etal-2022-flores}
Naman Goyal, Cynthia Gao, Vishrav Chaudhary, Peng-Jen Chen, Guillaume Wenzek,
  Da~Ju, Sanjana Krishnan, Marc{'}Aurelio Ranzato, Francisco Guzm{\'a}n, and
  Angela Fan. 2022.
\newblock \href {https://doi.org/10.1162/tacl_a_00474} {The {F}lores-101
  evaluation benchmark for low-resource and multilingual machine translation}.
\newblock \emph{Transactions of the Association for Computational Linguistics},
  10:522--538.

\bibitem[{Gu et~al.(2018)Gu, Hassan, Devlin, and Li}]{gu2018universal}
Jiatao Gu, Hany Hassan, Jacob Devlin, and Victor~O.K. Li. 2018.
\newblock Universal neural machine translation for extremely low resource
  languages.
\newblock In \emph{Proceedings of the Conference of the North {A}merican
  Chapter of the Association for Computational Linguistics: Human Language
  Technologies (NAACL-HLT)}.

\bibitem[{Hu et~al.(2021)Hu, Shen, Wallis, Allen-Zhu, Li, Wang, Wang, and
  Chen}]{hu2021lora}
Edward~J. Hu, Yelong Shen, Phillip Wallis, Zeyuan Allen-Zhu, Yuanzhi Li, Shean
  Wang, Lu~Wang, and Weizhu Chen. 2021.
\newblock \href {http://arxiv.org/abs/2106.09685} {Lora: Low-rank adaptation of
  large language models}.

\bibitem[{Ji et~al.(2023)Ji, Gong, Deng, Peng, Niu, Ma, and Li}]{ji2023better}
Yunjie Ji, Yan Gong, Yong Deng, Yiping Peng, Qiang Niu, Baochang Ma, and
  Xiangang Li. 2023.
\newblock \href {http://arxiv.org/abs/2304.07854} {Towards better instruction
  following language models for chinese: Investigating the impact of training
  data and evaluation}.

\bibitem[{Jiao et~al.(2023)Jiao, tse Huang, Wang, Wang, Shi, and
  Tu}]{jiao2023parrot}
Wenxiang Jiao, Jen tse Huang, Wenxuan Wang, Xing Wang, Shuming Shi, and
  Zhaopeng Tu. 2023.
\newblock Parrot: Translating during chat using large language models.
\newblock In \emph{ArXiv}.

\bibitem[{Johnson et~al.(2017)Johnson, Schuster, Le, Krikun, Wu, Chen, Thorat,
  Vi{\'e}gas, Wattenberg, Corrado et~al.}]{johnson2017google}
Melvin Johnson, Mike Schuster, Quoc~V Le, Maxim Krikun, Yonghui Wu, Zhifeng
  Chen, Nikhil Thorat, Fernanda Vi{\'e}gas, Martin Wattenberg, Greg Corrado,
  et~al. 2017.
\newblock Google’s multilingual neural machine translation system: Enabling
  zero-shot translation.
\newblock \emph{Transactions of the Association for Computational Linguistics},
  5:339--351.

\bibitem[{Kudo and Richardson(2018)}]{kudo-richardson-2018-sentencepiece}
Taku Kudo and John Richardson. 2018.
\newblock \href {https://doi.org/10.18653/v1/D18-2012} {{S}entence{P}iece: A
  simple and language independent subword tokenizer and detokenizer for neural
  text processing}.
\newblock In \emph{Proceedings of the 2018 Conference on Empirical Methods in
  Natural Language Processing: System Demonstrations}, pages 66--71, Brussels,
  Belgium. Association for Computational Linguistics.

\bibitem[{Lample and Conneau(2019)}]{lample2019crosslingual}
Guillaume Lample and Alexis Conneau. 2019.
\newblock \href {http://arxiv.org/abs/1901.07291} {Cross-lingual language model
  pretraining}.

\bibitem[{Li et~al.(2023)Li, Zhou, Huang, Cheng, and Chen}]{li2023eliciting}
Jiahuan Li, Hao Zhou, Shujian Huang, Shanbo Cheng, and Jiajun Chen. 2023.
\newblock \href {http://arxiv.org/abs/2305.15083} {Eliciting the translation
  ability of large language models via multilingual finetuning with translation
  instructions}.

\bibitem[{Liang et~al.(2023)Liang, Gonen, Mao, Hou, Goyal, Ghazvininejad,
  Zettlemoyer, and Khabsa}]{liang2023xlmv}
Davis Liang, Hila Gonen, Yuning Mao, Rui Hou, Naman Goyal, Marjan
  Ghazvininejad, Luke Zettlemoyer, and Madian Khabsa. 2023.
\newblock \href {http://arxiv.org/abs/2301.10472} {Xlm-v: Overcoming the
  vocabulary bottleneck in multilingual masked language models}.

\bibitem[{Libovický et~al.(2019)Libovický, Rosa, and
  Fraser}]{libovický2019languageneutral}
Jindřich Libovický, Rudolf Rosa, and Alexander Fraser. 2019.
\newblock \href {http://arxiv.org/abs/1911.03310} {How language-neutral is
  multilingual bert?}

\bibitem[{Lin et~al.(2022{\natexlab{a}})Lin, Mihaylov, Artetxe, Wang, Chen,
  Simig, Ott, Goyal, Bhosale, Du, Pasunuru, Shleifer, Koura, Chaudhary,
  O{'}Horo, Wang, Zettlemoyer, Kozareva, Diab, Stoyanov, and
  Li}]{lin-etal-2022-shot}
Xi~Victoria Lin, Todor Mihaylov, Mikel Artetxe, Tianlu Wang, Shuohui Chen,
  Daniel Simig, Myle Ott, Naman Goyal, Shruti Bhosale, Jingfei Du, Ramakanth
  Pasunuru, Sam Shleifer, Punit~Singh Koura, Vishrav Chaudhary, Brian O{'}Horo,
  Jeff Wang, Luke Zettlemoyer, Zornitsa Kozareva, Mona Diab, Veselin Stoyanov,
  and Xian Li. 2022{\natexlab{a}}.
\newblock \href {https://doi.org/10.18653/v1/2022.emnlp-main.616} {Few-shot
  learning with multilingual generative language models}.
\newblock In \emph{Proceedings of the 2022 Conference on Empirical Methods in
  Natural Language Processing}, pages 9019--9052, Abu Dhabi, United Arab
  Emirates. Association for Computational Linguistics.

\bibitem[{Lin et~al.(2022{\natexlab{b}})Lin, Mihaylov, Artetxe, Wang, Chen,
  Simig, Ott, Goyal, Bhosale, Du, Pasunuru, Shleifer, Koura, Chaudhary, O'Horo,
  Wang, Zettlemoyer, Kozareva, Diab, Stoyanov, and Li}]{lin2022fewshot}
Xi~Victoria Lin, Todor Mihaylov, Mikel Artetxe, Tianlu Wang, Shuohui Chen,
  Daniel Simig, Myle Ott, Naman Goyal, Shruti Bhosale, Jingfei Du, Ramakanth
  Pasunuru, Sam Shleifer, Punit~Singh Koura, Vishrav Chaudhary, Brian O'Horo,
  Jeff Wang, Luke Zettlemoyer, Zornitsa Kozareva, Mona Diab, Veselin Stoyanov,
  and Xian Li. 2022{\natexlab{b}}.
\newblock \href {http://arxiv.org/abs/2112.10668} {Few-shot learning with
  multilingual language models}.

\bibitem[{Lin et~al.(2019)Lin, Chen, Lee, Li, Zhang, Xia, Rijhwani, He, Zhang,
  Ma, Anastasopoulos, Littell, and Neubig}]{lin-etal-2019-choosing}
Yu-Hsiang Lin, Chian-Yu Chen, Jean Lee, Zirui Li, Yuyan Zhang, Mengzhou Xia,
  Shruti Rijhwani, Junxian He, Zhisong Zhang, Xuezhe Ma, Antonios
  Anastasopoulos, Patrick Littell, and Graham Neubig. 2019.
\newblock \href {https://doi.org/10.18653/v1/P19-1301} {Choosing transfer
  languages for cross-lingual learning}.
\newblock In \emph{Proceedings of the 57th Annual Meeting of the Association
  for Computational Linguistics}, pages 3125--3135, Florence, Italy.
  Association for Computational Linguistics.

\bibitem[{Muennighoff et~al.(2023)Muennighoff, Wang, Sutawika, Roberts,
  Biderman, Scao, Bari, Shen, Yong, Schoelkopf, Tang, Radev, Aji, Almubarak,
  Albanie, Alyafeai, Webson, Raff, and Raffel}]{muennighoff2023crosslingual}
Niklas Muennighoff, Thomas Wang, Lintang Sutawika, Adam Roberts, Stella
  Biderman, Teven~Le Scao, M~Saiful Bari, Sheng Shen, Zheng-Xin Yong, Hailey
  Schoelkopf, Xiangru Tang, Dragomir Radev, Alham~Fikri Aji, Khalid Almubarak,
  Samuel Albanie, Zaid Alyafeai, Albert Webson, Edward Raff, and Colin Raffel.
  2023.
\newblock \href {http://arxiv.org/abs/2211.01786} {Crosslingual generalization
  through multitask finetuning}.

\bibitem[{Muennighoff et~al.(2022)Muennighoff, Wang, Sutawika, Roberts,
  Biderman, Scao, Bari, Shen, Yong, Schoelkopf, Tang, Radev, Aji, Almubarak,
  Albanie, Alyafeai, Webson, Raff, and Raffel}]{Muennighoff2022CrosslingualGT}
Niklas Muennighoff, Thomas Wang, Lintang Sutawika, Adam Roberts, Stella~Rose
  Biderman, Teven~Le Scao, M~Saiful Bari, Sheng Shen, Zheng~Xin Yong, Hailey
  Schoelkopf, Xiangru Tang, Dragomir~R. Radev, Alham~Fikri Aji, Khalid
  Almubarak, Samuel Albanie, Zaid Alyafeai, Albert Webson, Edward Raff, and
  Colin Raffel. 2022.
\newblock Crosslingual generalization through multitask finetuning.
\newblock \emph{ArXiv}, abs/2211.01786.

\bibitem[{Neubig and Hu(2018)}]{neubig2018rapid}
Graham Neubig and Junjie Hu. 2018.
\newblock Rapid adaptation of neural machine translation to new languages.
\newblock In \emph{Proceedings of the Conference on Empirical Methods in
  Natural Language Processing (EMNLP)}.

\bibitem[{OpenAI(2023)}]{openai2023gpt4}
OpenAI. 2023.
\newblock \href {http://arxiv.org/abs/2303.08774} {Gpt-4 technical report}.

\bibitem[{Pires et~al.(2019)Pires, Schlinger, and
  Garrette}]{pires-etal-2019-multilingual}
Telmo Pires, Eva Schlinger, and Dan Garrette. 2019.
\newblock \href {https://doi.org/10.18653/v1/P19-1493} {How multilingual is
  multilingual {BERT}?}
\newblock In \emph{Proceedings of the 57th Annual Meeting of the Association
  for Computational Linguistics}, pages 4996--5001, Florence, Italy.
  Association for Computational Linguistics.

\bibitem[{Ponti et~al.(2020)Ponti, Glavaš, Majewska, Liu, Vulić, and
  Korhonen}]{ponti2020xcopa}
Edoardo~Maria Ponti, Goran Glavaš, Olga Majewska, Qianchu Liu, Ivan Vulić,
  and Anna Korhonen. 2020.
\newblock \href {http://arxiv.org/abs/2005.00333} {Xcopa: A multilingual
  dataset for causal commonsense reasoning}.

\bibitem[{Scao et~al.(2022)Scao, Fan, Akiki, Pavlick, Ili'c, Hesslow,
  Castagn'e, Luccioni, Yvon, Gall{\'e}, Tow, Rush, Biderman, Webson,
  Ammanamanchi, Wang, Sagot, Muennighoff, del Moral, Ruwase, Bawden, Bekman,
  McMillan-Major, Beltagy, Nguyen, Saulnier, Tan, Suarez, Sanh, Laurenccon,
  Jernite, Launay, Mitchell, Raffel, Gokaslan, Simhi, Etxabe, Aji, Alfassy,
  Rogers, Nitzav, Xu, Mou, Emezue, Klamm, Leong, van Strien, Adelani, Radev,
  Ponferrada, Levkovizh, Kim, Natan, Toni, Dupont, Kruszewski, Pistilli,
  ElSahar, Benyamina, Tran, Yu, Abdulmumin, Johnson, Gonzalez-Dios, de~la Rosa,
  Chim, Dodge, Zhu, Chang, Frohberg, Tobing, Bhattacharjee, Almubarak, Chen,
  Lo, von Werra, Weber, Phan, Allal, Tanguy, Dey, Mu{\~n}oz, Masoud, Grandury,
  vSavsko, Huang, Coavoux, Singh, Jiang, Vu, Jauhar, Ghaleb, Subramani,
  Kassner, Khamis, Nguyen, Espejel, de~Gibert, Villegas, Henderson, Colombo,
  Amuok, Lhoest, Harliman, Bommasani, L'opez, Ribeiro, Osei, Pyysalo, Nagel,
  Bose, Muhammad, Sharma, Longpre, Nikpoor, Silberberg, Pai, Zink, Torrent,
  Schick, Thrush, Danchev, Nikoulina, Laippala, Lepercq, Prabhu, Alyafeai,
  Talat, Raja, Heinzerling, Si, Salesky, Mielke, Lee, Sharma, Santilli,
  Chaffin, Stiegler, Datta, Szczechla, Chhablani, Wang, Pandey, Strobelt,
  Fries, Rozen, Gao, Sutawika, Bari, Al-shaibani, Manica, Nayak, Teehan,
  Albanie, Shen, Ben-David, Bach, Kim, Bers, F{\'e}vry, Neeraj, Thakker,
  Raunak, Tang, Yong, Sun, Brody, Uri, Tojarieh, Roberts, Chung, Tae, Phang,
  Press, Li, Narayanan, Bourfoune, Casper, Rasley, Ryabinin, Mishra, Zhang,
  Shoeybi, Peyrounette, Patry, Tazi, Sanseviero, von Platen, Cornette,
  Lavall'ee, Lacroix, Rajbhandari, Gandhi, Smith, Requena, Patil, Dettmers,
  Baruwa, Singh, Cheveleva, Ligozat, Subramonian, N'ev'eol, Lovering, Garrette,
  Tunuguntla, Reiter, Taktasheva, Voloshina, Bogdanov, Winata, Schoelkopf,
  Kalo, Novikova, Forde, Tang, Kasai, Kawamura, Hazan, Carpuat, Clinciu, Kim,
  Cheng, Serikov, Antverg, van~der Wal, Zhang, Zhang, Gehrmann, Pais, Shavrina,
  Scialom, Yun, Limisiewicz, Rieser, Protasov, Mikhailov, Pruksachatkun,
  Belinkov, Bamberger, Kasner, Rueda, Pestana, Feizpour, Khan, Faranak, Santos,
  Hevia, Unldreaj, Aghagol, Abdollahi, Tammour, HajiHosseini, Behroozi,
  Ajibade, Saxena, Ferrandis, Contractor, Lansky, David, Kiela, Nguyen, Tan,
  Baylor, Ozoani, Mirza, Ononiwu, Rezanejad, Jones, Bhattacharya, Solaiman,
  Sedenko, Nejadgholi, Passmore, Seltzer, Sanz, Fort, Dutra, Samagaio, Elbadri,
  Mieskes, Gerchick, Akinlolu, McKenna, Qiu, Ghauri, Burynok, Abrar, Rajani,
  Elkott, Fahmy, Samuel, An, Kromann, Hao, Alizadeh, Shubber, Wang, Roy,
  Viguier, Le, Oyebade, Le, Yang, Nguyen, Kashyap, Palasciano, Callahan,
  Shukla, Miranda-Escalada, Singh, Beilharz, Wang, de~Brito, Zhou, Jain, Xu,
  Fourrier, Perin'an, Molano, Yu, Manjavacas, Barth, Fuhrimann, Altay, Bayrak,
  Burns, Vrabec, Bello, Dash, Kang, Giorgi, Golde, Posada, Sivaraman,
  Bulchandani, Liu, Shinzato, de~Bykhovetz, Takeuchi, P{\`a}mies, Castillo,
  Nezhurina, Sanger, Samwald, Cullan, Weinberg, Wolf, Mihaljcic, Liu, Freidank,
  Kang, Seelam, Dahlberg, Broad, Muellner, Fung, Haller, Chandrasekhar,
  Eisenberg, Martin, Canalli, Su, Su, Cahyawijaya, Garda, Deshmukh, Mishra,
  Kiblawi, Ott, Sang-aroonsiri, Kumar, Schweter, Bharati, Laud, Gigant,
  Kainuma, Kusa, Labrak, Bajaj, Venkatraman, Xu, Xu, chao Xu, Tan, Xie, Ye,
  Bras, Belkada, and Wolf}]{Scao2022BLOOMA1}
Teven~Le Scao, Angela Fan, Christopher Akiki, Elizabeth-Jane Pavlick, Suzana
  Ili'c, Daniel Hesslow, Roman Castagn'e, Alexandra~Sasha Luccioni, Franccois
  Yvon, Matthias Gall{\'e}, Jonathan Tow, Alexander~M. Rush, Stella~Rose
  Biderman, Albert Webson, Pawan~Sasanka Ammanamanchi, Thomas Wang, Beno{\^i}t
  Sagot, Niklas Muennighoff, Albert~Villanova del Moral, Olatunji Ruwase,
  Rachel Bawden, Stas Bekman, Angelina McMillan-Major, Iz~Beltagy, Huu Nguyen,
  Lucile Saulnier, Samson Tan, Pedro~Ortiz Suarez, Victor Sanh, Hugo
  Laurenccon, Yacine Jernite, Julien Launay, Margaret Mitchell, Colin Raffel,
  Aaron Gokaslan, Adi Simhi, Aitor~Soroa Etxabe, Alham~Fikri Aji, Amit Alfassy,
  Anna Rogers, Ariel~Kreisberg Nitzav, Canwen Xu, Chenghao Mou, Chris~C.
  Emezue, Christopher Klamm, Colin Leong, Daniel~Alexander van Strien,
  David~Ifeoluwa Adelani, Dragomir~R. Radev, Eduardo~Gonz'alez Ponferrada,
  Efrat Levkovizh, Ethan Kim, Eyal~Bar Natan, Francesco~De Toni, G{\'e}rard
  Dupont, Germ{\'a}n Kruszewski, Giada Pistilli, Hady ElSahar, Hamza Benyamina,
  Hieu~Trung Tran, Ian Yu, Idris Abdulmumin, Isaac Johnson, Itziar
  Gonzalez-Dios, Javier de~la Rosa, Jenny Chim, Jesse Dodge, Jian Zhu, Jonathan
  Chang, Jorg Frohberg, Josephine~L. Tobing, Joydeep Bhattacharjee, Khalid
  Almubarak, Kimbo Chen, Kyle Lo, Leandro von Werra, Leon Weber, Long Phan,
  Loubna~Ben Allal, Ludovic Tanguy, Manan Dey, Manuel~Romero Mu{\~n}oz, Maraim
  Masoud, Mar'ia Grandury, Mario vSavsko, Max Huang, Maximin Coavoux, Mayank
  Singh, Mike Tian-Jian Jiang, Minh~Chien Vu, Mohammad~Ali Jauhar, Mustafa
  Ghaleb, Nishant Subramani, Nora Kassner, Nurulaqilla Khamis, Olivier Nguyen,
  Omar Espejel, Ona de~Gibert, Paulo Villegas, Peter Henderson, Pierre Colombo,
  Priscilla Amuok, Quentin Lhoest, Rheza Harliman, Rishi Bommasani, Roberto
  L'opez, Rui Ribeiro, Salomey Osei, Sampo Pyysalo, Sebastian Nagel, Shamik
  Bose, Shamsuddeen~Hassan Muhammad, Shanya Sharma, S.~Longpre, Somaieh
  Nikpoor, Stanislav Silberberg, Suhas Pai, Sydney Zink, Tiago~Timponi Torrent,
  Timo Schick, Tristan Thrush, Valentin Danchev, Vassilina Nikoulina, Veronika
  Laippala, Violette Lepercq, Vrinda Prabhu, Zaid Alyafeai, Zeerak Talat, Arun
  Raja, Benjamin Heinzerling, Chenglei Si, Elizabeth Salesky, Sabrina~J.
  Mielke, Wilson~Y. Lee, Abheesht Sharma, Andrea Santilli, Antoine Chaffin,
  Arnaud Stiegler, Debajyoti Datta, Eliza Szczechla, Gunjan Chhablani, Han
  Wang, Harshit Pandey, Hendrik Strobelt, Jason~Alan Fries, Jos Rozen, Leo Gao,
  Lintang Sutawika, M~Saiful Bari, Maged~S. Al-shaibani, Matteo Manica,
  Nihal~V. Nayak, Ryan Teehan, Samuel Albanie, Sheng Shen, Srulik Ben-David,
  Stephen~H. Bach, Taewoon Kim, Tali Bers, Thibault F{\'e}vry, Trishala Neeraj,
  Urmish Thakker, Vikas Raunak, Xiang Tang, Zheng~Xin Yong, Zhiqing Sun, Shaked
  Brody, Y~Uri, Hadar Tojarieh, Adam Roberts, Hyung~Won Chung, Jaesung Tae,
  Jason Phang, Ofir Press, Conglong Li, Deepak Narayanan, Hatim Bourfoune,
  Jared Casper, Jeff Rasley, Max Ryabinin, Mayank Mishra, Minjia Zhang,
  Mohammad Shoeybi, Myriam Peyrounette, Nicolas Patry, Nouamane Tazi, Omar
  Sanseviero, Patrick von Platen, Pierre Cornette, Pierre~Franccois Lavall'ee,
  R{\'e}mi Lacroix, Samyam Rajbhandari, Sanchit Gandhi, Shaden Smith,
  St{\'e}phane Requena, Suraj Patil, Tim Dettmers, Ahmed Baruwa, Amanpreet
  Singh, Anastasia Cheveleva, Anne-Laure Ligozat, Arjun Subramonian, Aur'elie
  N'ev'eol, Charles Lovering, Daniel~H Garrette, Deepak~R. Tunuguntla, Ehud
  Reiter, Ekaterina Taktasheva, Ekaterina Voloshina, Eli Bogdanov, Genta~Indra
  Winata, Hailey Schoelkopf, Jan-Christoph Kalo, Jekaterina Novikova,
  Jessica~Zosa Forde, Xiangru Tang, Jungo Kasai, Ken Kawamura, Liam Hazan,
  Marine Carpuat, Miruna Clinciu, Najoung Kim, Newton Cheng, Oleg Serikov, Omer
  Antverg, Oskar van~der Wal, Rui Zhang, Ruochen Zhang, Sebastian Gehrmann,
  S.~Osher Pais, Tatiana Shavrina, Thomas Scialom, Tian Yun, Tomasz
  Limisiewicz, Verena Rieser, Vitaly Protasov, Vladislav Mikhailov, Yada
  Pruksachatkun, Yonatan Belinkov, Zachary Bamberger, Zdenvek Kasner, Alice
  Rueda, Amanda Pestana, Amir Feizpour, Ammar Khan, Amy Faranak, Ananda
  Santa~Rosa Santos, Anthony Hevia, Antigona Unldreaj, Arash Aghagol, Arezoo
  Abdollahi, Aycha Tammour, Azadeh HajiHosseini, Bahareh Behroozi,
  Benjamin~Olusola Ajibade, Bharat~Kumar Saxena, Carlos~Mu{\~n}oz Ferrandis,
  Danish Contractor, David~M. Lansky, Davis David, Douwe Kiela, Duong~Anh
  Nguyen, Edward Tan, Emily Baylor, Ezinwanne Ozoani, Fatim~T Mirza, Frankline
  Ononiwu, Habib Rezanejad, H.A. Jones, Indrani Bhattacharya, Irene Solaiman,
  Irina Sedenko, Isar Nejadgholi, Jan Passmore, Joshua Seltzer, Julio~Bonis
  Sanz, Karen Fort, L{\'i}via~Macedo Dutra, Mairon Samagaio, Maraim Elbadri,
  Margot Mieskes, Marissa Gerchick, Martha Akinlolu, Michael McKenna, Mike Qiu,
  M.~K.~K. Ghauri, Mykola Burynok, Nafis Abrar, Nazneen Rajani, Nour Elkott,
  Nourhan Fahmy, Olanrewaju Samuel, Ran An, R.~P. Kromann, Ryan Hao, Samira
  Alizadeh, Sarmad Shubber, Silas~L. Wang, Sourav Roy, Sylvain Viguier,
  Thanh-Cong Le, Tobi Oyebade, Trieu Nguyen~Hai Le, Yoyo Yang, Zachary~Kyle
  Nguyen, Abhinav~Ramesh Kashyap, Alfredo Palasciano, Alison Callahan, Anima
  Shukla, Antonio Miranda-Escalada, Ayush~Kumar Singh, Benjamin Beilharz,
  Bo~Wang, Caio Matheus~Fonseca de~Brito, Chenxi Zhou, Chirag Jain, Chuxin Xu,
  Cl{\'e}mentine Fourrier, Daniel~Le'on Perin'an, Daniel Molano, Dian Yu,
  Enrique Manjavacas, Fabio Barth, Florian Fuhrimann, Gabriel Altay, Giyaseddin
  Bayrak, Gully~A. Burns, Helena~U. Vrabec, Iman~I.B. Bello, Isha Dash, Ji~Soo
  Kang, John Giorgi, Jonas Golde, Jose~David Posada, Karthi Sivaraman, Lokesh
  Bulchandani, Lu~Liu, Luisa Shinzato, Madeleine~Hahn de~Bykhovetz, Maiko
  Takeuchi, Marc P{\`a}mies, Mar{\'i}a~Andrea Castillo, Marianna Nezhurina,
  Mario Sanger, Matthias Samwald, Michael Cullan, Michael Weinberg, M~Wolf,
  Mina Mihaljcic, Minna Liu, Moritz Freidank, Myungsun Kang, Natasha Seelam,
  Nathan Dahlberg, Nicholas~Michio Broad, Nikolaus Muellner, Pascale Fung,
  Patricia Haller, R.~Chandrasekhar, R.~Eisenberg, Robert Martin, Rodrigo~L.
  Canalli, Rosaline Su, Ruisi Su, Samuel Cahyawijaya, Samuele Garda, Shlok~S
  Deshmukh, Shubhanshu Mishra, Sid Kiblawi, Simon Ott, Sinee Sang-aroonsiri,
  Srishti Kumar, Stefan Schweter, Sushil~Pratap Bharati, T.~A. Laud, Th'eo
  Gigant, Tomoya Kainuma, Wojciech Kusa, Yanis Labrak, Yashasvi Bajaj,
  Y.~Venkatraman, Yifan Xu, Ying Xu, Yun chao Xu, Zhee~Xao Tan, Zhongli Xie,
  Zifan Ye, Mathilde Bras, Younes Belkada, and Thomas Wolf. 2022.
\newblock Bloom: A 176b-parameter open-access multilingual language model.
\newblock \emph{ArXiv}, abs/2211.05100.

\bibitem[{Schwenk et~al.(2021)Schwenk, Chaudhary, Sun, Gong, and
  Guzm{\'a}n}]{schwenk-etal-2021-wikimatrix}
Holger Schwenk, Vishrav Chaudhary, Shuo Sun, Hongyu Gong, and Francisco
  Guzm{\'a}n. 2021.
\newblock \href {https://doi.org/10.18653/v1/2021.eacl-main.115}
  {{W}iki{M}atrix: Mining 135{M} parallel sentences in 1620 language pairs from
  {W}ikipedia}.
\newblock In \emph{Proceedings of the 16th Conference of the European Chapter
  of the Association for Computational Linguistics: Main Volume}, pages
  1351--1361, Online. Association for Computational Linguistics.

\bibitem[{Sennrich et~al.(2016)Sennrich, Haddow, and
  Birch}]{sennrich-etal-2016-neural}
Rico Sennrich, Barry Haddow, and Alexandra Birch. 2016.
\newblock \href {https://doi.org/10.18653/v1/P16-1162} {Neural machine
  translation of rare words with subword units}.
\newblock In \emph{Proceedings of the 54th Annual Meeting of the Association
  for Computational Linguistics (Volume 1: Long Papers)}, pages 1715--1725,
  Berlin, Germany. Association for Computational Linguistics.

\bibitem[{Shi et~al.(2022)Shi, Suzgun, Freitag, Wang, Srivats, Vosoughi, Chung,
  Tay, Ruder, Zhou, Das, and Wei}]{shi2022language}
Freda Shi, Mirac Suzgun, Markus Freitag, Xuezhi Wang, Suraj Srivats, Soroush
  Vosoughi, Hyung~Won Chung, Yi~Tay, Sebastian Ruder, Denny Zhou, Dipanjan Das,
  and Jason Wei. 2022.
\newblock \href {http://arxiv.org/abs/2210.03057} {Language models are
  multilingual chain-of-thought reasoners}.

\bibitem[{Tai et~al.(2020)Tai, Kung, Dong, Comiter, and
  Kuo}]{tai-etal-2020-exbert}
Wen Tai, H.~T. Kung, Xin Dong, Marcus Comiter, and Chang-Fu Kuo. 2020.
\newblock \href {https://doi.org/10.18653/v1/2020.findings-emnlp.129}
  {ex{BERT}: Extending pre-trained models with domain-specific vocabulary under
  constrained training resources}.
\newblock In \emph{Findings of the Association for Computational Linguistics:
  EMNLP 2020}, pages 1433--1439, Online. Association for Computational
  Linguistics.

\bibitem[{Tiedemann(2012)}]{tiedemann-2012-parallel}
J{\"o}rg Tiedemann. 2012.
\newblock \href
  {http://www.lrec-conf.org/proceedings/lrec2012/pdf/463_Paper.pdf} {Parallel
  data, tools and interfaces in {OPUS}}.
\newblock In \emph{Proceedings of the Eighth International Conference on
  Language Resources and Evaluation ({LREC}'12)}, pages 2214--2218, Istanbul,
  Turkey. European Language Resources Association (ELRA).

\bibitem[{Touvron et~al.(2023{\natexlab{a}})Touvron, Lavril, Izacard, Martinet,
  Lachaux, Lacroix, Rozière, Goyal, Hambro, Azhar, Rodriguez, Joulin, Grave,
  and Lample}]{llama1}
Hugo Touvron, Thibaut Lavril, Gautier Izacard, Xavier Martinet, Marie-Anne
  Lachaux, Timothée Lacroix, Baptiste Rozière, Naman Goyal, Eric Hambro,
  Faisal Azhar, Aurelien Rodriguez, Armand Joulin, Edouard Grave, and Guillaume
  Lample. 2023{\natexlab{a}}.
\newblock \href {http://arxiv.org/abs/2302.13971} {Llama: Open and efficient
  foundation language models}.

\bibitem[{Touvron et~al.(2023{\natexlab{b}})Touvron, Martin, Stone, Albert,
  Almahairi, Babaei, Bashlykov, Batra, Bhargava, Bhosale et~al.}]{llama2}
Hugo Touvron, Louis Martin, Kevin Stone, Peter Albert, Amjad Almahairi, Yasmine
  Babaei, Nikolay Bashlykov, Soumya Batra, Prajjwal Bhargava, Shruti Bhosale,
  et~al. 2023{\natexlab{b}}.
\newblock Llama 2: Open foundation and fine-tuned chat models.
\newblock \emph{arXiv preprint arXiv:2307.09288}.

\bibitem[{Wang et~al.(2019)Wang, Cho, and Gu}]{bbpe}
Changhan Wang, Kyunghyun Cho, and Jiatao Gu. 2019.
\newblock \href {http://arxiv.org/abs/1909.03341} {Neural machine translation
  with byte-level subwords}.

\bibitem[{Wei et~al.(2023)Wei, Wei, Lin, Li, Zhang, Ren, Li, Wan, Cao, Xie, Hu,
  Li, Hui, Yu, Liu, Yang, Huang, and Xie}]{wei2023polylm}
Xiangpeng Wei, Haoran Wei, Huan Lin, Tianhao Li, Pei Zhang, Xingzhang Ren, Mei
  Li, Yu~Wan, Zhiwei Cao, Binbin Xie, Tianxiang Hu, Shangjie Li, Binyuan Hui,
  Bowen Yu, Dayiheng Liu, Baosong Yang, Fei Huang, and Jun Xie. 2023.
\newblock \href {http://arxiv.org/abs/2307.06018} {Polylm: An open source
  polyglot large language model}.

\bibitem[{Wu and Dredze(2020)}]{wu-dredze-2020-languages}
Shijie Wu and Mark Dredze. 2020.
\newblock \href {https://doi.org/10.18653/v1/2020.repl4nlp-1.16} {Are all
  languages created equal in multilingual {BERT}?}
\newblock In \emph{Proceedings of the 5th Workshop on Representation Learning
  for NLP}, pages 120--130, Online. Association for Computational Linguistics.

\bibitem[{Xue et~al.(2021)Xue, Constant, Roberts, Kale, Al-Rfou, Siddhant,
  Barua, and Raffel}]{xue-etal-2021-mt5}
Linting Xue, Noah Constant, Adam Roberts, Mihir Kale, Rami Al-Rfou, Aditya
  Siddhant, Aditya Barua, and Colin Raffel. 2021.
\newblock \href {https://doi.org/10.18653/v1/2021.naacl-main.41} {m{T}5: A
  massively multilingual pre-trained text-to-text transformer}.
\newblock In \emph{Proceedings of the 2021 Conference of the North American
  Chapter of the Association for Computational Linguistics: Human Language
  Technologies}, pages 483--498, Online. Association for Computational
  Linguistics.

\bibitem[{Yang et~al.(2023)Yang, Li, Zhang, and Zong}]{yang2023bigtrans}
Wen Yang, Chong Li, Jiajun Zhang, and Chengqing Zong. 2023.
\newblock Bigtrans: Augmenting large language models with multilingual
  translation capability over 100 languages.
\newblock \emph{arXiv preprint arXiv:2305.18098}.

\bibitem[{Yang et~al.(2019)Yang, Zhang, Tar, and Baldridge}]{yang2019pawsx}
Yinfei Yang, Yuan Zhang, Chris Tar, and Jason Baldridge. 2019.
\newblock \href {http://arxiv.org/abs/1908.11828} {Paws-x: A cross-lingual
  adversarial dataset for paraphrase identification}.

\bibitem[{Ye et~al.(2018)Ye, Devendra, Matthieu, Sarguna, and
  Graham}]{Ye2018WordEmbeddings}
Qi~Ye, Sachan Devendra, Felix Matthieu, Padmanabhan Sarguna, and Neubig Graham.
  2018.
\newblock When and why are pre-trained word embeddings useful for neural
  machine translation.
\newblock In \emph{HLT-NAACL}.

\bibitem[{Yong et~al.(2023)Yong, Zhang, Forde, Wang, Subramonian, Lovenia,
  Cahyawijaya, Winata, Sutawika, Cruz, Tan, Phan, Garcia, Solorio, and
  Aji}]{yong2023prompting}
Zheng-Xin Yong, Ruochen Zhang, Jessica~Zosa Forde, Skyler Wang, Arjun
  Subramonian, Holy Lovenia, Samuel Cahyawijaya, Genta~Indra Winata, Lintang
  Sutawika, Jan Christian~Blaise Cruz, Yin~Lin Tan, Long Phan, Rowena Garcia,
  Thamar Solorio, and Alham~Fikri Aji. 2023.
\newblock \href {http://arxiv.org/abs/2303.13592} {Prompting multilingual large
  language models to generate code-mixed texts: The case of south east asian
  languages}.

\bibitem[{Yuan et~al.(2023)Yuan, Lu, Zhu, Kong, Li, Qiao, and Xu}]{legoMT}
Fei Yuan, Yinquan Lu, Wenhao Zhu, Lingpeng Kong, Lei Li, Yu~Qiao, and Jingjing
  Xu. 2023.
\newblock \href {https://doi.org/10.18653/v1/2023.findings-acl.731}
  {{L}ego-{MT}: Learning detachable models for massively multilingual machine
  translation}.
\newblock In \emph{Findings of the Association for Computational Linguistics:
  ACL 2023}, pages 11518--11533, Toronto, Canada. Association for Computational
  Linguistics.

\bibitem[{Zhang et~al.(2020)Zhang, Williams, Titov, and
  Sennrich}]{zhang2020improving}
Biao Zhang, Philip Williams, Ivan Titov, and Rico Sennrich. 2020.
\newblock Improving massively multilingual neural machine translation and
  zero-shot translation.
\newblock \emph{arXiv preprint arXiv:2004.11867}.

\bibitem[{Zhang et~al.(2022)Zhang, Roller, Goyal, Artetxe, Chen, Chen, Dewan,
  Diab, Li, Lin, Mihaylov, Ott, Shleifer, Shuster, Simig, Koura, Sridhar, Wang,
  and Zettlemoyer}]{zhang2022opt}
Susan Zhang, Stephen Roller, Naman Goyal, Mikel Artetxe, Moya Chen, Shuohui
  Chen, Christopher Dewan, Mona Diab, Xian Li, Xi~Victoria Lin, Todor Mihaylov,
  Myle Ott, Sam Shleifer, Kurt Shuster, Daniel Simig, Punit~Singh Koura, Anjali
  Sridhar, Tianlu Wang, and Luke Zettlemoyer. 2022.
\newblock \href {http://arxiv.org/abs/2205.01068} {Opt: Open pre-trained
  transformer language models}.

\bibitem[{Zhu et~al.(2023{\natexlab{a}})Zhu, Liu, Dong, Xu, Huang, Kong, Chen,
  and Li}]{zhu2023multilingual}
Wenhao Zhu, Hongyi Liu, Qingxiu Dong, Jingjing Xu, shujian Huang, Lingpeng
  Kong, Jiajun Chen, and Lei Li. 2023{\natexlab{a}}.
\newblock Multilingual machine translation with large language models:
  Empirical results and analysis.
\newblock \emph{arXiv preprint arXiv:2304.04675}.

\bibitem[{Zhu et~al.(2023{\natexlab{b}})Zhu, Lv, Dong, Yuan, Xu, Huang, Kong,
  Chen, and Li}]{zhu2023extrapolating}
Wenhao Zhu, Yunzhe Lv, Qingxiu Dong, Fei Yuan, Jingjing Xu, Shujian Huang,
  Lingpeng Kong, Jiajun Chen, and Lei Li. 2023{\natexlab{b}}.
\newblock \href {http://arxiv.org/abs/2308.04948} {Extrapolating large language
  models to non-english by aligning languages}.

\end{thebibliography}
\bibliographystyle{acl_natbib}

\clearpage
\label{sec:appendix}

\newpage

\appendix

\section{Instruction Tuning}
\label{appendix:it}

Instruction Tuning is a method used to train large language models to follow specific instructions to solve a task. It's a form of supervised learning where the model is trained on a dataset consisting of pairs of instructions and corresponding outputs.

Table~\ref{tab:it-case} shows a simple example in the Lego-MT dataset, presented in the format used for instruction tuning:

\begin{table}[h]
    \centering 
    \footnotesize
    \begin{tabular}{|p{7.2cm}|}
        \hline
        Below is an instruction that describes a task, paired with an input that provides further context. Write a response that appropriately completes the request. \\
        \textbf{Instruction:} Translate the following sentences from English to French. \\
        \textbf{Input:} Dogs are the main source of transmission of rabies to humans. \\
        \textbf{Response:} Les chiens sont la principale source de transmission de la rage. \\
        \hline
    \end{tabular}
    \caption{Instruction tuning case based on Lego-MT dataset.}
    \label{tab:it-case}
\end{table}

\begin{table*}[h]
    \centering
    \resizebox{1\textwidth}{!}{
    \begin{tabular}{c|c|c|c|c|c|c|c|c|c}
        \toprule
        \textbf{Family-1} &\textbf{Family-2} & \textbf{Family-3} & \textbf{ISO} & \textbf{Language} & \textbf{Lang Family-1} & \textbf{Family-2} & \textbf{Family-3} & \textbf{ISO} & \textbf{Language} \\ 
        \midrule
        \multirow{50}{*}{Indo-European} & Armenian & ~ & hy & Armenian & Kartvelian & Karto-Zan & Georgian & ka & Georgian \\ 
        \cline{2-5}  \cline{6-10}
         & \multirow{14}{*}{Balto-Slavic} & \multirow{2}{*}{Baltic} & lt & Lithuanian & Koreanic & Korean & ~ & ko & Korean \\ 
         \cline{6-10}
         &  &  & lv & Latvian & \multirow{2}{*}{Kra–Dai} & \multirow{2}{*}{Tai} & \multirow{2}{*}{Southwestern Tai} & lo & Lao \\ 
         \cline{3-5}
         &  & \multirow{12}{*}{Slavic} & be & Belarusian &  &  &  & \cellcolor{gray!25} th & \cellcolor{gray!25} Thai \\ 
         \cline{6-10}
         &  &  & bg & Bulgarian & Mongolic & Central & Mongolian & mn & Mongolian \\ 
         \cline{6-10}
         &  &  & bs & Bosnian & \multirow{13}{*}{Niger–Congo} & \multirow{13}{*}{Atlantic–Congo} & Atlantic & wo & Wolof \\ 
         \cline{8-10}
         &  &  & cs & Czech & &  & \multirow{2}{*}{Benue–Congo} & ln & Lingala \\ 
         &  &  & hr & Croatian & & &  & \cellcolor{gray!25} ns & \cellcolor{gray!25} Northern Sotho \\ 
         \cline{8-10}
         &  &  & mk & Macedonian & &  & \multirow{10}{*}{Volta-Congo} & lg & Luganda \\ 
         &  &  & pl & Polish &  &  &  & ny & Nyanja \\ 
         &  &  & ru & Russian & &  &  & sn & Shona \\ 
         &  &  & sk & Slovak & & &  & sw & Swahili \\ 
         &  &  & sl & Slovenian &  &  &  & umb & Umbundu \\ 
         &  &  & sr & Serbian &  &  & & xh & Xhosa \\ 
         &  &  & uk & Ukrainian &  &  &  & yo & Yoruba \\
         \cline{2-5}
         & \multirow{2}{*}{Celtic} & \multirow{2}{*}{Insular Celtic} & cy & Welsh &  &  &  & zu & Zulu \\ 
         &  & & ga & Irish &  &  &  & ig & Igbo \\ 
         \cline{2-5}
         & \multirow{9}{*}{Germanic} & \multirow{2}{*}{North Germanic} & is & Icelandic &  &  &  & kam & Kamba \\ 
         \cline{8-10}
         &  & & sv & Swedish &  &  & West Atlantic & ff & Fulani \\ 
         \cline{3-5}          \cline{6-10}
         &  & \multirow{2}{*}{Northwest Germanic} & da & Danish & Nilo-Saharan & Eastern & Nilotic & \cellcolor{gray!25} luo & \cellcolor{gray!25} Dholuo \\ 
         \cline{6-10}
         &  &  & no & Norwegian & Portuguese & Afro-Portuguese & Upper Guinea Creole & kea & Kabuverdianu \\ 
         \cline{3-5} \cline{6-10}
         &  & \multirow{5}{*}{West Germanic} & \cellcolor{gray!25} af & \cellcolor{gray!25} Afrikaans & \multirow{3}{*}{Sino-Tibetan} & \multirow{2}{*}{Sinitic} & \multirow{2}{*}{Chinese} & \cellcolor{gray!25} zh & \cellcolor{gray!25} Chinese \\ 
         &  & & de & German &  & &  & zhtrad & Chinese \\ 
         \cline{7-10}
         &  & & en & English & & Tibeto-Burman & Lolo-Burmese & my & Burmese \\
         \cline{6-10}
         &  & & lb & Luxembourgish & \multirow{5}{*}{Turkic} & \multirow{5}{*}{Common} & Karluk & uz & Uzbek \\ 
         \cline{8-10}
         &  & & nl & Dutch &  &  & \multirow{2}{*}{Kipchak} & kk & Kazakh \\ 
         \cline{2-5}
         & Graeco-Phrygian & Hellenic & el & Greek &  &  &  & ky & Kyrgyz \\ 
         \cline{2-5} \cline{8-10}
         & Indo-Aryan & Eastern & bn & Bengali &  &  & \multirow{2}{*}{Oghuz} & az & Azerbaijani \\ 
         \cline{2-5}
         & \multirow{13}{*}{Indo-Iranian} & \multirow{9}{*}{Indo-Aryan} & as & Assamese &  &  &  & \cellcolor{gray!25} tr & \cellcolor{gray!25} Turkish \\ 
         \cline{6-10}
         &  & & gu & Gujarati & \multirow{3}{*}{Uralic} & Finno-Permic & Finno-Samic & \cellcolor{gray!25} et & \cellcolor{gray!25} Estonian \\ 
         \cline{7-10}
         &  & & hi & Hindi &  & \multirow{2}{*}{Finno-Ugric} & Finnic & fi & Finnish \\ 
         \cline{8-10}
         &  & & mr & Marathi &  &  & Ugric & hu & Hungarian \\ 
         \cline{6-10}
         &  & & ne & Nepali & \multirow{7}{*}{Afro-Asiatic} & Chadic & West Chadic & \cellcolor{gray!25} ha & \cellcolor{gray!25} Hausa \\ 
         \cline{7-10}
         &  & & or & Odia &  & \multirow{2}{*}{Cushitic} & \multirow{2}{*}{Lowland East Cushitic} & om & Oromo \\ 
         &  & & pa & Punjabi &  &  &  & so & Somali \\ 
         \cline{7-10}
         &  & & sd & Sindhi &  & \multirow{4}{*}{Semitic} & \multirow{4}{*}{West Semitic} & am & Amharic \\ 
         &  & & ur & Urdu &  &  & & ar & Arabic \\ 
         \cline{3-5}
         &  & \multirow{4}{*}{Iranian} & fa & Persian &  &  &  & \cellcolor{gray!25} he & \cellcolor{gray!25} Hebrew \\ 
         &  &  & ku & Kurdish &  &  & & mt & Maltese \\ 
        \cline{6-10}
         &  &  & ps & Pashto & \multirow{2}{*}{Austroasiatic} & Khmer & ~ & km & Khmer \\ 
         \cline{7-10}
         &  &  & tg & Tajik &  & Vietic & Viet–Muong & vi & Vietnamese \\ 
         \cline{2-5}         \cline{6-10}
         & \multirow{9}{*}{Italic} & \multirow{9}{*}{Latino-Faliscan} & ast & Asturian & \multirow{6}{*}{Austronesian} & \multirow{6}{*}{Malayo-Polynesian} & Javanese & jv & Javanese \\ 
         \cline{8-10}
         &  &  & ca & Catalan &  &  & \multirow{2}{*}{Malayic} & id & Indonesian \\ 
         &  &  & es & Spanish &  &  &  & ms & Malay \\ 
         \cline{8-10}
         &  &  & fr & French &  &  & Oceanic & \cellcolor{gray!25} mi & \cellcolor{gray!25} Maori \\ 
         \cline{8-10}
         &  &  & gl & Galician &  &  & \multirow{2}{*}{Philippine} & ceb & Cebuano \\ 
         &  &  & it & Italian &  &  &  & tl & Tagalog \\ 
         \cline{6-10}
         &  &  & oc & Occitan & \multirow{4}{*}{Dravidian} & South-Central & Telugu & te & Telugu \\ 
         \cline{7-10}
         &  &  & pt & Portuguese &  & \multirow{3}{*}{Southern} & \multirow{3}{*}{Tamil–Kannada} & kn & Kannada \\ 
         &  &  & \cellcolor{gray!25} ro & \cellcolor{gray!25} Romanian &  &  &  & ml & Malayalam \\ 
         \cline{1-5}
        Japonic & Japanese & ~ & ja & Japanese &  &  &  & \cellcolor{gray!25} ta & \cellcolor{gray!25} Tamil \\ 
    \bottomrule
    \end{tabular}}
    \caption{This table provides information on the language families of all 101 languages included in FLores-101. The language family information is presented at three levels, denoted as “Lang Family-x”, where ‘x’ stands for the level (1, 2, or 3). For ease of reference, we also provide the ISO code and the full name of each language. Languages that are used in the inherent multilingual analysis are highlighted with a gray background.}
    \label{table:language_family_info}
\end{table*}

\section{BBPE}
\label{appendix:bbpe}
In a multilingual context encompassing a diverse range of scripts, the base vocabulary comprising subwords can become exceedingly large, leading to inefficiency and sparsity. To mitigate this problem, BBPE has emerged as the standard practice in most modern language modeling efforts~\citep{Muennighoff2022CrosslingualGT,Scao2022BLOOMA1,zhang2022opt,llama1,llama2}, which leverages UTF-8 encoding that encodes each Unicode character into 1 to 4 one-byte (8-bit) code units. BBPE is a tokenization algorithm capable of tokenizing any word in any language, thereby eliminating the necessity for an unknown token. It optimizes vocabulary sharing across numerous languages and delivers superior performance, facilitating knowledge transfer between languages with non-overlapping character sets.

\section{Language Information}
\label{appendix:lang_info}

In this section, we classify languages according to their respective language families, as depicted in Table~\ref{table:language_family_info}. We standardize all language codes using the ISO 639-1 standard. For clarity, we list all languages by their full names and shade the corresponding languages in gray for easy identification.

\section{Hyper-parameter Setting}
\label{appendix:hyper_selection}

\begin{figure}[!h]
    \centering
    \includegraphics[width=0.8\linewidth]{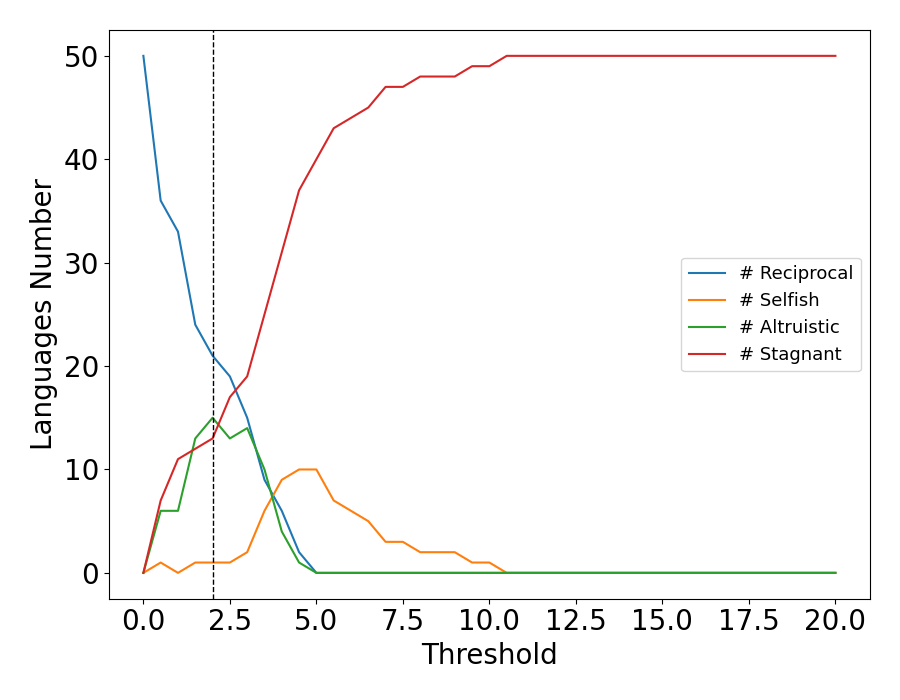}
    \caption{Hyper-parameter setting. ``Threshold'' refers to the significant changes before and after tuning, which are calculated by dividing the performance after tuning by the performance before the tuning. ``\# Reciprocal'' denotes the count of languages in the Reciprocal quadrant. The experimental result demonstrates that a substantial increase in the threshold value could lead to all languages being classified into the Stagnant quadrant.}
    \label{fig:hyper_selection}
\end{figure}

We use the criteria to measure the bilingual/multilingual performance changes before and after fine-tuning:
\begin{equation*}
        \Delta_{\mathrm{lg}} = 
    \begin{cases}
        \frac{P_{\text{post}} }{P_{\text{pre}}} - 2,& \text{if } P_{\text{pre}}\geq T\\
         \frac{P_{\text{post}} - 2T}{P_{\text{pre}}},              & \text{otherwise}
    \end{cases}
\end{equation*}
The threshold term, $T$, is used to smooth the dramatic numerical change that might be caused by low-performing languages(e.g., performance change from 0.01 to 0.02, although negligible, will be considered significant without re-balancing using $T$). We set $T$ to the vanilla model’s average translation performance on the Flores-101 dataset. 

The hyper-parameter, set to a value of 2, defines the thresholds for determining significant changes before and after tuning. Here, we consider a language to have significant bilingual/multilingual performance changes if the performance after tuning is twice that of the performance before tuning. In Figure~\ref{fig:hyper_selection}, we have thoroughly tested different significance thresholds and found that if we consider a 20-fold difference (a very large value) in performance before and after tuning, then all languages would be regarded as stagnant languages.

\section{Quadrant Division}
\label{appendix:appendix_division}

We use some different publicly multilingual datasets: Lego-MT, Wikimatrix \& Newcommentary, and Ted, which come from a different domain, as shown in Table~\ref{tab:data_stat}.

\begin{table}[!h]
    \centering
    \footnotesize
    \resizebox{0.95\linewidth}{!}{
    \begin{tabular}{lcl}
    \toprule
    \textbf{Dataset} & \textbf{\# Language} & \textbf{Domain} \\
    \midrule
    Lego-MT &  100 & Web\\
    Wikimatrix \&  & \multirow{2}{*}{50} & \multirow{2}{*}{Wikipedia and News}\\
    Newscommentary & &  \\
    Ted & 55 &  TED talk \\
    \bottomrule
    \end{tabular}
    }
    \caption{Statistics of various publicly accessible parallel multilingual corpora.}
    \label{tab:data_stat}
\end{table}

Conducting a comprehensive evaluation of the translation performance for all en$\rightarrow$x pairs in Flores-101 across all models is a task that demands significant labor and resources. Therefore, we randomly select one representative language from each language family for subsequent testing, as shown in Table~\ref{tab:representaive_lang}. 

\begin{table}[!h]
    \centering
    \footnotesize
    \resizebox{0.95\linewidth}{!}{
    \begin{tabular}{cc|cc}
    \toprule
    \textbf{Lang} & \textbf{Language Family} & \textbf{Lang} & \textbf{Language Family} \\
    \midrule
ha & Afro-Asiatic & he & Afro-Asiatic \\
mi & Austronesian & ta & Dravidian \\
af & Indo-European & ro & Indo-European \\
th & Kra–Dai & ns & Niger–Congo \\
luo & Nilo-Saharan & zh & Sino-Tibetan \\
tr & Turkic & et & Uralic \\
    \bottomrule
    \end{tabular}
    }
    \caption{Representative languages information. Within the Indo-European language family, we choose to include af in addition to ro, which is a first language in South Africa and not initially listed as a supported language by LLaMA.}
    \label{tab:representaive_lang}
\end{table}

The bilingual and multilingual performance of the model trained on the TED dataset on Flores-101 devtest is shown in Table~\ref{tab:all_res}. 
\begin{table*}[!h]
    \centering
    \footnotesize
    \resizebox{0.95\linewidth}{!}{
    \begin{tabular}{c|ccccccc|cc|ccc|c}
    \toprule
\textbf{LG} & \textbf{en$\rightarrow$mi} & \textbf{en$\rightarrow$luo} & \textbf{en$\rightarrow$ns} & \textbf{en$\rightarrow$ha} & \textbf{en$\rightarrow$ta} & \textbf{en$\rightarrow$tr} & \textbf{en$\rightarrow$he} & \textbf{en$\rightarrow$af} & \textbf{en$\rightarrow$ro} & \textbf{en$\rightarrow$th} & \textbf{en$\rightarrow$zh} & \textbf{en$\rightarrow$et} & \textbf{en$\rightarrow$LG}  \\ 
\midrule
LLaMA & 2.3 & 3.1 & 3.3 & 3.1 & 0.4 & 2.4 & 0.5 & 3.5 & 3.6 & 0.8 & 0.5 & 1.6 & -\\
\midrule
ar & 0.9 & 1.5 & 0.8 & 0.3 & 0.4 & 1.9 & 2.5 & 7.8 & 16.1 & 0.3 & 1.4 & 1.4 & 2.2  \\
hy & 1.0 & 0.9 & 0.6 & 0.0 & 0.3 & 1.9 & 1.2 & 3.7 & 4.3 & 0.1 & 0.9 & 1.1 & 0.9  \\
az & 1.2 & 2.1 & 1.9 & 1.3 & 0.0 & 1.8 & 0.0 & 2.5 & 0.1 & 0.0 & 0.5 & 1.1 & 0.0  \\
be & 0.9 & 2.4 & 1.9 & 2.1 & 0.0 & 1.3 & 0.0 & 2.0 & 0.0 & 0.0 & 0.5 & 0.9 & 0.0  \\
bn & 0.6 & 2.0 & 1.4 & 1.5 & 0.0 & 0.9 & 0.0 & 1.8 & 0.0 & 0.0 & 0.5 & 0.7 & 0.1  \\
bs & 1.1 & 2.2 & 1.9 & 1.7 & 0.0 & 1.4 & 0.0 & 1.8 & 0.0 & 0.0 & 0.4 & 0.9 & 0.4  \\
bg & 1.5 & 1.7 & 1.3 & 0.5 & 0.3 & 2.1 & 2.0 & 8.2 & 12.1 & 0.2 & 1.1 & 1.5 & 18.8  \\
my & 0.3 & 0.3 & 0.4 & 0.0 & 0.2 & 1.5 & 1.1 & 3.0 & 0.9 & 0.1 & 1.0 & 0.8 & 0.0  \\
zh & 0.8 & 1.7 & 1.4 & 1.3 & 0.0 & 1.7 & 0.0 & 1.7 & 0.1 & 0.0 & 0.5 & 0.9 & 0.5  \\
hr & 2.0 & 2.6 & 2.2 & 1.8 & 0.4 & 2.6 & 2.8 & 10.4 & 19.2 & 0.5 & 1.5 & 1.9 & 13.8  \\
cs & 1.6 & 2.1 & 1.3 & 0.9 & 0.4 & 2.1 & 2.0 & 9.1 & 14.4 & 0.3 & 1.1 & 1.7 & 16.5  \\
da & 2.2 & 2.7 & 2.3 & 2.1 & 0.4 & 2.6 & 2.6 & 10.8 & 18.2 & 0.5 & 1.3 & 1.9 & 24.5  \\
nl & 1.6 & 2.3 & 1.6 & 1.3 & 0.4 & 2.0 & 2.1 & 10.4 & 15.8 & 0.4 & 1.3 & 1.7 & 23.1  \\
et & 1.3 & 2.0 & 1.7 & 1.1 & 0.4 & 2.5 & 2.6 & 8.7 & 16.9 & 0.4 & 1.3 & 1.9 & 1.9  \\
fi & 1.5 & 2.4 & 2.0 & 1.6 & 0.4 & 2.3 & 2.2 & 8.0 & 15.7 & 0.3 & 1.2 & 1.7 & 2.0  \\
fr & 2.2 & 2.9 & 2.7 & 2.3 & 0.4 & 2.9 & 2.6 & 9.5 & 18.4 & 0.5 & 1.6 & 1.8 & 36.8  \\
gl & 1.7 & 2.3 & 2.0 & 0.7 & 0.4 & 2.7 & 2.4 & 8.9 & 14.9 & 0.3 & 1.1 & 1.8 & 3.1  \\
ka & 0.9 & 0.7 & 0.3 & 0.0 & 0.4 & 1.7 & 1.8 & 6.7 & 11.8 & 0.2 & 1.2 & 1.4 & 0.1  \\
de & 1.8 & 2.4 & 1.9 & 1.6 & 0.4 & 2.2 & 2.4 & 10.1 & 16.7 & 0.5 & 1.6 & 1.8 & 25.9  \\
el & 1.4 & 2.2 & 1.7 & 0.9 & 0.4 & 2.3 & 2.8 & 11.2 & 21.4 & 0.3 & 1.8 & 1.7 & 5.4  \\
he & 1.7 & 2.4 & 2.0 & 1.1 & 0.4 & 2.6 & 3.4 & 8.2 & 21.4 & 0.5 & 2.3 & 1.8 & 3.4  \\
hi & 0.3 & 0.4 & 0.2 & 0.0 & 0.3 & 1.4 & 1.6 & 3.6 & 5.8 & 0.1 & 1.2 & 0.8 & 4.1  \\
hu & 1.6 & 2.3 & 1.6 & 1.1 & 0.4 & 2.2 & 1.6 & 8.4 & 14.2 & 0.3 & 1.1 & 1.6 & 6.4  \\
id & 2.4 & 3.1 & 2.9 & 2.5 & 0.4 & 3.0 & 2.9 & 9.1 & 19.9 & 0.6 & 1.4 & 1.8 & 7.3  \\
it & 2.2 & 2.7 & 2.5 & 2.0 & 0.4 & 2.7 & 2.4 & 9.7 & 18.6 & 0.5 & 1.5 & 2.0 & 23.8  \\
ja & 1.6 & 2.0 & 1.8 & 1.1 & 0.4 & 1.9 & 3.0 & 6.7 & 19.2 & 0.5 & 2.3 & 1.6 & 5.4  \\
kk & 1.2 & 2.8 & 2.6 & 2.6 & 0.1 & 1.9 & 0.1 & 2.9 & 0.6 & 0.3 & 0.5 & 1.3 & 0.3  \\
ko & 1.0 & 1.9 & 1.4 & 1.1 & 0.4 & 1.9 & 1.8 & 7.4 & 17.3 & 0.3 & 1.9 & 1.5 & 2.9  \\
lt & 1.6 & 2.3 & 2.1 & 1.6 & 0.4 & 2.5 & 2.4 & 8.9 & 19.6 & 0.5 & 1.4 & 1.9 & 1.0  \\
mk & 1.1 & 0.2 & 0.2 & 0.0 & 0.3 & 1.7 & 1.9 & 7.7 & 9.8 & 0.2 & 1.2 & 1.2 & 4.4  \\
ms & 1.3 & 2.6 & 2.1 & 1.9 & 0.1 & 1.7 & 0.0 & 2.5 & 0.1 & 0.0 & 0.5 & 1.1 & 2.7  \\
mr & 0.8 & 2.5 & 2.5 & 1.3 & 0.3 & 2.2 & 2.2 & 5.8 & 9.4 & 0.4 & 1.1 & 1.4 & 1.1  \\
mn & 1.1 & 1.5 & 1.2 & 0.1 & 0.2 & 2.0 & 0.7 & 4.3 & 2.3 & 0.1 & 0.9 & 1.1 & 0.0  \\
fa & 0.7 & 0.9 & 0.3 & 0.1 & 0.4 & 1.6 & 1.8 & 7.3 & 15.9 & 0.2 & 1.7 & 1.3 & 2.6  \\
pl & 1.8 & 2.4 & 2.1 & 1.7 & 0.4 & 2.3 & 2.3 & 9.2 & 15.0 & 0.4 & 1.4 & 1.7 & 12.6  \\
pt & 2.0 & 2.6 & 2.2 & 1.9 & 0.3 & 2.6 & 2.5 & 11.6 & 20.7 & 0.5 & 1.7 & 1.9 & 36.1  \\
ro & 1.8 & 2.3 & 1.6 & 1.1 & 0.4 & 2.1 & 2.1 & 8.9 & 15.5 & 0.4 & 1.2 & 1.6 & 15.5  \\
ru & 1.3 & 2.0 & 1.3 & 0.7 & 0.4 & 1.9 & 1.5 & 7.2 & 9.7 & 0.3 & 1.1 & 1.5 & 16.6  \\
sr & 1.8 & 2.2 & 1.7 & 1.3 & 0.4 & 2.3 & 2.6 & 10.3 & 17.7 & 0.5 & 1.3 & 1.7 & 2.0  \\
sk & 1.9 & 2.3 & 2.1 & 1.5 & 0.4 & 2.5 & 2.5 & 9.0 & 16.9 & 0.4 & 1.2 & 1.9 & 5.7  \\
sl & 1.8 & 2.4 & 1.9 & 1.4 & 0.3 & 2.3 & 2.3 & 8.3 & 16.1 & 0.4 & 1.3 & 1.8 & 8.0  \\
ku & 0.4 & 1.2 & 1.5 & 0.4 & 0.3 & 1.6 & 2.0 & 4.5 & 6.1 & 0.3 & 1.2 & 1.2 & 0.0  \\
es & 2.2 & 2.8 & 2.5 & 2.1 & 0.4 & 2.8 & 2.8 & 11.3 & 20.8 & 0.6 & 1.8 & 1.9 & 24.8  \\
sv & 1.9 & 2.6 & 2.1 & 1.9 & 0.4 & 2.3 & 2.6 & 10.4 & 18.2 & 0.5 & 1.3 & 1.8 & 24.7  \\
ta & 1.3 & 2.9 & 2.4 & 2.1 & 0.1 & 2.1 & 0.0 & 2.9 & 0.4 & 0.0 & 0.6 & 1.5 & 0.1  \\
th & 0.7 & 0.8 & 0.7 & 0.1 & 0.4 & 2.2 & 1.8 & 4.5 & 13.2 & 0.4 & 1.5 & 1.3 & 0.4  \\
tr & 1.7 & 2.3 & 1.8 & 1.4 & 0.4 & 2.4 & 2.3 & 8.2 & 17.1 & 0.4 & 1.4 & 1.7 & 2.4  \\
uk & 1.3 & 1.9 & 1.4 & 0.8 & 0.4 & 1.8 & 1.4 & 7.0 & 8.8 & 0.2 & 0.9 & 1.3 & 3.0  \\
ur & 0.6 & 0.9 & 0.5 & 0.3 & 0.0 & 1.0 & 0.0 & 0.7 & 0.0 & 0.0 & 0.6 & 0.4 & 0.0  \\
vi & 1.8 & 2.7 & 2.4 & 1.9 & 0.3 & 2.5 & 2.6 & 8.5 & 15.6 & 0.5 & 1.2 & 1.8 & 2.6  \\
    \bottomrule
    \end{tabular}
    }
    \caption{Assessing the bilingual and multilingual capabilities: a performance evaluation of the model trained on the TED dataset across all representative languages using the Flores-101 devtest. The experimental results show the significant improvement in multilingual performance embodied in the en$\rightarrow$af and en$\rightarrow$ro.  }
    \label{tab:all_res}
\end{table*}

\section{Stagnant Quadrant}
\label{appendix:over-tokenization}

The LLaMA tokenizer, built on the BBPE algorithm, serves as the foundation for multilingual language processing tasks. Its universal applicability across all languages, coupled with the elimination of the need for an “unknown” token, enhances vocabulary sharing and boosts its robustness. However, a phenomenon known as over-tokenization, marked by excessive segmentation of text into tokens, may occur in certain languages, which could potentially affect the efficiency of language processing tasks.

To thoroughly examine the “over-tokenization”, we conduct our research using the MC4~\citep{xue-etal-2021-mt5} and Flores-101~\citep{goyal-etal-2022-flores} dataset. Despite having only 1012 samples, Flores-101 provides a high-quality multilingual parallel corpus that allows for an in-depth exploration of the variations in expressing the same sentence across different languages.

\begin{figure*}[t]
   \begin{subfigure}{1\textwidth}
	\centering
		\includegraphics[width=1\textwidth]{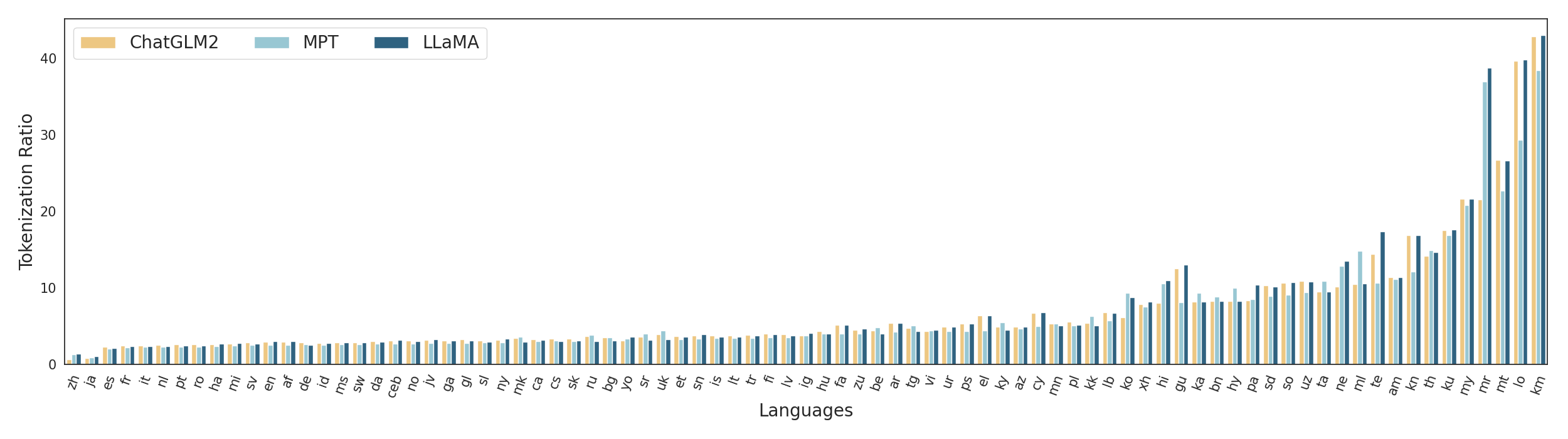}
		\caption{Tokenization analysis on MC4 dataset.}
  	\label{fig:mc4_all_tokenization}
	\end{subfigure}%
    \hfill
	\begin{subfigure}{1\textwidth}
	\centering
		\includegraphics[width=1\textwidth]{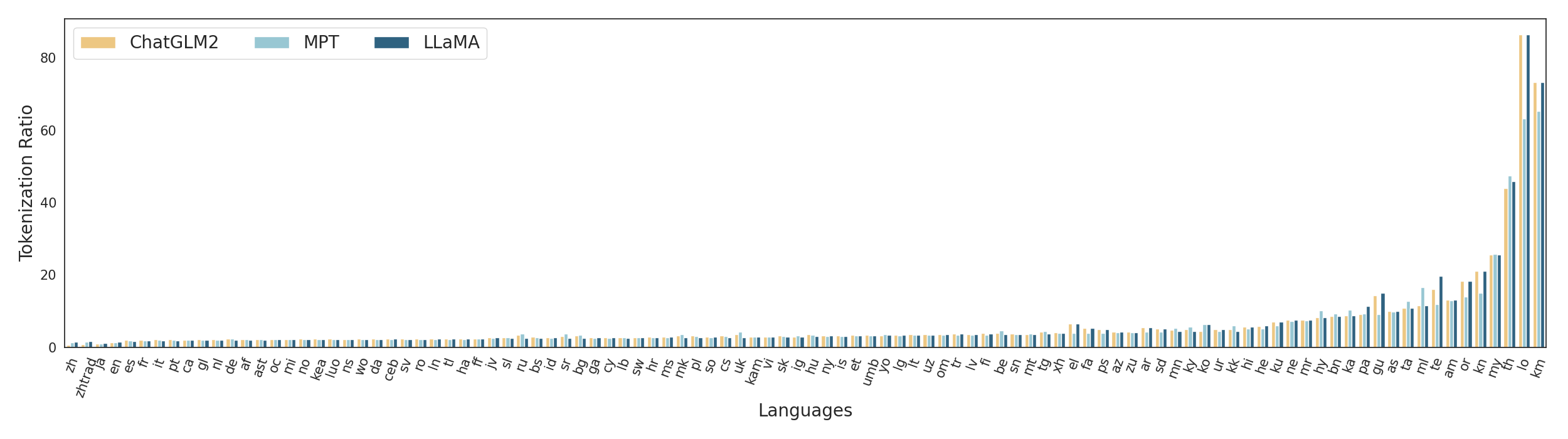}
		\caption{Tokenization analysis on Flores-101 dataset.}
  	\label{fig:flores_all_tokenization}
	\end{subfigure}%
\caption{An over-tokenization phenomenon in low-resource languages across different datasets and LLMs. The tokenization ratios of LLaMA, ChatGLM2, and MPT are calculated by dividing the length of the tokenized sequence by the sentence length. For space-separated languages, the sentence length is measured by the number of words, while for other languages it is measured by the number of characters. The length of the tokenized sequence refers to the number of tokens obtained after applying the tokenizer. Languages characterized by over-tokenization will exhibit this trait across various LLMs.}
\label{fig:all_tokenization}
\end{figure*}

The over-tokenization phenomenon is observable across various datasets and LLMs. For certain languages, such as te and lo, the length of the tokenized sequence that LLaMA processes can extend to 300 or even more. Interestingly, analysis results from the Flores-101 dataset reveal that languages prone to over-tokenization require more tokens to express the same meaning. The magnitude of this phenomenon is notably larger than what was observed in the MC4 dataset, as shown in Figure~\ref{fig:all_tokenization}.

We also present tuning results based on our analysis of the Flores-101 dataset, where we examined the effects of full bilingual fine-tuning and Lora tuning on varying amounts of data, as shown in Table~\ref{tab:training_over_tokenized_lgs_flores}. Interestingly, we found that the characteristics of stagnant language are preserved.

\begin{table*}[!h]
    \centering
    \footnotesize
    \resizebox{0.8\linewidth}{!}{
    \begin{tabular}{c|c|c|ccccc|ccccc}
    \toprule
    \multirow{2}{*}{\textbf{Setting}} & \multirow{2}{*}{\textbf{Ratio}} & \multirow{2}{*}{\textbf{LLaMA} } & \multicolumn{5}{c|}{\textbf{Full Bilingual Fine-Tuning}} &  \multicolumn{5}{c}{\textbf{LoRA Bilingual Tuning}} \\
   & & & 10k & 20k & 40K & 80k & 160k & 10k & 20k & 40K & 80k & 160k \\ 
    \midrule
en$\rightarrow$es & 1.7 & 4.8 & 23.5 & 23.8 & 25.2 & 23.6 & 25.9 & 26.4 & 25.8 & 26.6 & 26.3 & 26.9	  \\
en$\rightarrow$pt & 1.9 & 6.0 & 41.3 & 41.1 & 41.3 & 40.6 & 39.7 & 42.0 & 42.0 & 42.4 & 42.0 & 41.6	  \\
en$\rightarrow$ca & 1.9 & 5.7 & 34.9 & 35.7 & 37.0 & 38.5 & 39.2 & 37.3 & 37.7 & 38.1 & 38.6 & 39.2	  \\
en$\rightarrow$de & 2.0 & 4.7 & 22.5 & 24.8 & 25.9 & 30.8 & 31.2 & 27.8 & 26.8 & 27.3 & 31.9 & 32.6	  \\
en$\rightarrow$no & 2.2 & 3.2 & 21.2 & 22.2 & 24.0 & 25.6 & 28.4 & 19.6 & 20.1 & 21.0 & 22.1 & 24.0	  \\
en$\rightarrow$ro & 2.3 & 3.5 & 28.3 & 28.7 & 29.6 & 30.8 & 34.3 & 29.8 & 30.0 & 30.9 & 31.2 & 32.7	  \\
en$\rightarrow$da & 2.3 & 4.9 & 31.9 & 32.2 & 32.8 & 34.8 & 36.4 & 33.4 & 34.0 & 34.5 & 35.3 & 36.1	  \\
en$\rightarrow$bs & 2.6 & 2.0 & 23.2 & 25.2 & 26.5 & 28.5 & 30.0 & 21.7 & 22.8 & 24.2 & 25.0 & 25.2	  \\
\midrule
en$\rightarrow$as & 10.0 & 0.2 & 3.2 & 4.7 & 6.8 & 8.2 & 9.6 & 0.5 & 0.6 & 0.9 & 1.4 & 2.2	  \\
en$\rightarrow$ta & 11.0 & 0.4 & 2.2 & 4.3 & 9.6 & 15.3 & 21.4 & 0.4 & 0.6 & 1.0 & 1.9 & 3.4	  \\
en$\rightarrow$pa & 11.4 & 0.3 & 2.3 & 4.2 & 6.8 & 9.7 & 14.5 & 0.4 & 0.8 & 1.2 & 1.7 & 2.7	 \\
en$\rightarrow$ml & 11.6 & 0.2 & 3.1 & 7.4 & 13.5 & 20.3 & 22.5 & 0.6 & 0.9 & 1.7 & 3.3 & 4.1	  \\
en$\rightarrow$am & 13.1 & 0.2 & 1.3 & 4.6 & 9.6 & 14.5 & 18.2 & 0.1 & 0.1 & 0.2 & 0.4 & 1.1	  \\
en$\rightarrow$gu & 15.0 & 0.3 & 2.3 & 2.2 & 4.4 & 10.0 & 13.2 & 1.0 & 1.1 & 1.5 & 1.9 & 3.1	  \\
en$\rightarrow$or & 18.4 & 0.3 & 0.9 & 1.6 & 1.6 & 1.0 & 0.8 & 0.3 & 0.5 & 0.5 & 0.1 & 0.1	  \\
en$\rightarrow$te & 19.7 & 0.7 & 4.2 & 8.2 & 12.8 & 17.3 & 20.3 & 0.6 & 0.8 & 1.7 & 2.9 & 5.3	  \\
en$\rightarrow$kn & 21.1 & 0.3 & 1.0 & 1.5 & 3.0 & 5.6 & 9.9 & 0.5 & 0.4 & 0.5 & 0.8 & 1.0	  \\
en$\rightarrow$my & 25.7 & 0.3 & 1.0 & 2.0 & 4.1 & 7.3 & 9.4 & 0.1 & 0.1 & 0.3 & 0.3 & 0.4	  \\
en$\rightarrow$th & 45.9 & 0.8 & 2.6 & 4.0 & 6.0 & 8.4 & 12.3 & 1.2 & 1.5 & 1.9 & 2.8 & 4.1	  \\
en$\rightarrow$km & 73.3 & 1.1 & 1.6 & 3.1 & 6.2 & 10.1 & 13.4 & 0.2 & 0.2 & 0.5 & 0.9 & 1.5	  \\
en$\rightarrow$lo & 86.5 & 1.8 & 1.5 & 2.3 & 3.7 & 7.1 & 9.8 & 0.4 & 0.7 & 0.7 & 0.6 & 0.7	  \\
    \bottomrule
    \end{tabular}
    }
    \caption{This refers to the relationship between stagnant languages and the characteristic of over-tokenization. Here, the "Ratio" is defined as the number of tokens in a sequence after applying the tokenizer, divided by the sentence length. The sentence length is measured by the number of words for space-separated languages and characters for others.}
    \label{tab:training_over_tokenized_lgs_flores}
\end{table*}

\section{Single-layer Tuning}
\label{appendix:single-layer}

To determine whether fine-tuning parameters of layers other than the embedding layer in the model is equally effective, we conducted a bilingual translation task on eight language pairs in the Flores101 dataset. These models were fine-tuned on the Alpaca-En dataset, which was primarily used as the training set to minimize any potential impact from language variations. The results of these tests are displayed in Table~\ref{tab:full-part-ft}. In these tests, English served as the source language, while the target languages comprised eight different languages.

As observed from the table, the average scores of fine-tuning the embedding layer and Layer 0 are the highest, and they are very close to each other. The model’s performance gradually decreases as the layer number increases, with a noticeable drop around the middle layers (Layers 15-17). This trend is remarkably consistent across all language pair tests.

The aforementioned results suggest that solely fine-tuning the parameters of the lower layers can also activate the model’s multilingual capabilities, and its effectiveness is comparable to that of embedding fine-tuning. Furthermore, the activation of different language capabilities in the model through single-layer fine-tuning occurs synchronously.

Additionally, we fine-tuned all the lower layers, from Layer 0 to Layer 14, together. As shown in Table~\ref{tab:training_all_low_layers}, this strategy did not yield any additional gains compared to the other tuning strategies.

\begin{table*}[!h]
    \centering
    \footnotesize
    \resizebox{0.8\textwidth}{!}{
    \begin{tabular}{l|cccccccc|c}
        \toprule
        \textbf{Setting} & \textbf{en$\rightarrow$ro} & \textbf{en$\rightarrow$es} & \textbf{en$\rightarrow$de} & \textbf{en$\rightarrow$ca} & \textbf{en$\rightarrow$pt} & \textbf{en$\rightarrow$da}  &  \textbf{en$\rightarrow$no}& \textbf{en$\rightarrow$bs} & \textbf{AVG.} \\
        \midrule
   FT & 27.1 & 23.5 & 24.5 & 34.3 & 40.5 & \textbf{32.3} & \textbf{20.9} & \textbf{22.4} & 28.2 \\
   LoRA  & 28.8 & 26.6 & 30.3 & \textbf{36.6} & 40.3 & 31.5 & 18.2 & 20.3 & 29.1\\
   Embed FT & \textbf{29.1} & \textbf{26.8} & \textbf{31.0} & 35.9 & \textbf{41.1} & 32.1 & 18.3 & 19.4 & \textbf{29.2} \\ 
        \midrule
Layer 0  & \textbf{29.2}  & 26.6  & \textbf{30.9}  & \textbf{37.2}  & \textbf{41.5}  & \textbf{32.4}  & 18.7  & 20.8  & \textbf{29.7} \\
Layer 1  & 28.9  & 26.2  & 30.1  & 36.2  & 40.5  & 32.1  & 19.0  & 20.3  & 29.2 \\
Layer 2  & 28.9  & 26.7  & 30.6  & 36.4  & 40.6  & 32.2  & 18.9  & \textbf{21.4}  & 29.5 \\
Layer 3  & 28.8  & 26.6  & 30.4  & 36.6  & 40.6  & 31.8  & 18.6  & 20.6  & 29.2 \\
Layer 4 & 29.0  & 26.8  & 30.4  & 36.7  & 40.5  & 32.1  & 18.9  & 20.4  & 29.3 \\
Layer 5  & 28.9  & \textbf{27.0}  & 30.9  & 37.1  & 41.3  & 32.1  & 19.0  & 20.8  & 29.6 \\
Layer 6  & 29.0  & 26.8  & 30.5  & 36.7  & 40.8  & 31.5  & 19.0  & 20.3  & 29.3 \\
Layer 7  & 28.7  & 26.4  & 30.7  & 36.1  & 40.3  & 32.0  & 18.7  & 19.6  & 29.1 \\
Layer 8  & 29.1  & 26.2  & 30.0  & 36.4  & 40.4  & 31.6  & \textbf{19.2}  & 19.7  & 29.1 \\
Layer 9 & 28.8  & 26.3  & 30.2  & 35.8  & 40.2  & 31.6  & 19.1  & 19.5  & 28.9 \\
Layer 10  & 27.8  & 25.8  & 29.7  & 35.5  & 39.9  & 30.8  & 18.7  & 16.1  & 28.0 \\
Layer 11  & 28.0  & 25.6  & 29.9  & 35.5  & 39.4  & 30.9  & 18.8  & 17.1  & 28.2 \\
Layer 12  & 27.9  & 25.5  & 29.2  & 34.8  & 38.2  & 30.6  & 17.2  & 15.4  & 27.4 \\
Layer 13  & 27.8  & 25.6  & 29.1  & 34.1  & 38.3  & 30.4  & 17.3  & 16.5  & 27.4 \\
Layer 14  & 25.1  & 24.7  & 28.5  & 32.1  & 36.2  & 29.4  & 15.8  & 10.1  & 25.2 \\
\midrule
Layer 15  & 15.7  & 22.6  & 25.4  & 27.2  & 27.7  & 24.2  & 11.2  & 2.5  & 19.6 \\
Layer 16  & 15.2  & 20.3  & 23.2  & 26.5  & 18.9  & 20.2  & 10.4  & 3.2  & 17.2 \\
Layer 17  & 19.0  & 21.0  & 23.1  & 23.6  & 22.1  & 20.2  & 11.1  & 5.0  & 18.1 \\
Layer 18  & 7.1  & 6.7  & 8.9  & 7.5  & 5.8  & 10.1  & 5.6  & 3.1  & 6.8 \\
Layer 19  & 6.2  & 4.0  & 6.4  & 3.0  & 4.5  & 4.7  & 3.9  & 1.7  & 4.3 \\
Layer 20  & 6.1  & 5.4  & 4.0  & 3.9  & 6.0  & 5.9  & 4.7  & 2.5  & 4.8 \\
Layer 21  & 5.0  & 5.0  & 3.2  & 2.5  & 4.2  & 5.1  & 3.9  & 2.2  & 3.9 \\
Layer 22  & 5.4  & 5.3  & 2.9  & 3.7  & 6.6  & 7.7  & 3.9  & 2.6  & 4.8 \\
Layer 23  & 4.2  & 2.6  & 0.8  & 1.4  & 2.8  & 6.1  & 3.2  & 1.7  & 2.9 \\
Layer 24  & 4.3  & 3.5  & 2.9  & 1.8  & 5.2  & 5.1  & 3.4  & 2.1  & 3.5 \\
Layer 25  & 4.7  & 2.7  & 2.0  & 1.9  & 7.7  & 6.3  & 3.1  & 2.0  & 3.8 \\
Layer 26  & 4.7  & 2.7  & 3.8  & 2.2  & 6.3  & 4.7  & 3.0  & 2.4  & 3.7 \\
Layer 27  & 5.1  & 1.3  & 4.4  & 2.5  & 6.3  & 5.6  & 4.6  & 2.3  & 4.0 \\
Layer 28  & 4.6  & 1.6  & 4.3  & 2.7  & 4.9  & 3.8  & 3.3  & 2.6  & 3.5 \\
Layer 29  & 4.1  & 2.9  & 5.2  & 4.3  & 6.7  & 6.8  & 3.6  & 2.9  & 4.6 \\
Layer 30  & 4.8  & 2.6  & 5.6  & 4.2  & 6.1  & 5.3  & 4.1  & 2.8  & 4.4 \\
Layer 31  & 4.3  & 2.8  & 3.8  & 4.2  & 4.6  & 6.3  & 3.9  & 2.8  & 4.1 \\
        \bottomrule
    \end{tabular}
    }
    \caption{Single-layer fine-tuning results on Alpaca-En dataset. The layers of the LLaMA-7B model, excluding the embedding layer, are numbered according to their distance from the embedding layer, with the closest being Layer 0 and the furthest being Layer 31. The term ``+ Layer $i$'' indicates that only the $i$ th layer is fine-tuned, with the other parts of parameters fixed.}
\label{tab:full-part-ft}
\end{table*}

\begin{table*}[!h]
    \centering
    \footnotesize
    \resizebox{0.95\textwidth}{!}{
    \begin{tabular}{c|ccccccccccc|c}
\toprule
 \textbf{Size} & \textbf{en$\rightarrow$da} & \textbf{en$\rightarrow$ca} & \textbf{en$\rightarrow$cs} & \textbf{en$\rightarrow$bg} & \textbf{en$\rightarrow$pl} & \textbf{en$\rightarrow$es} & \textbf{en$\rightarrow$fr} & \textbf{en$\rightarrow$de} & \textbf{en$\rightarrow$pt} & \textbf{en$\rightarrow$ru} & \textbf{en$\rightarrow$nl} & \textbf{AVG.} \\ 
\midrule
\rowcolor{lightgray!30}     \multicolumn{13}{c}{\textbf{Bilingual Full Fine-Tuning}} \\
\midrule
       10k & 31.9 & 34.9 & 23.9 & 26.0 & 17.0 & 23.5 & 32.5 & 22.5 & 41.3 & 24.3 & 18.7 & 27.0 \\ 
       20k & 32.2 & 35.8 & 24.5 & 26.5 & 18.4 & 23.8 & 31.7 & 24.8 & 41.1 & 24.2 & 18.9 & 27.4 \\ 
       40k & 32.8 & 37.0 & 25.4 & 27.4 & 18.8 & 25.2 & 34.1 & 25.9 & 41.3 & 24.1 & 22.1 & 28.6 \\ 
       160k & 36.4 & 39.2 & 27.1 & 31.8 & 19.7 & 25.9 & 39.1 & 31.2 & 39.7 & 24.6 & 24.3 & 30.8 \\ 
\midrule
\rowcolor{lightgray!30}     \multicolumn{13}{c}{\textbf{Bilingual Embedding Fine-Tuning}} \\
\midrule
        10k & 26.4 & 30.1 & 16.6 & 19.6 & 12.6 & 23.7 & 34.7 & 23.1 & 33.3 & 19.1 & 21.2 & 23.7 \\ 
        20k & 33.1 & 37.3 & 24.4 & 26.5 & 18.6 & 26.4 & 41.1 & 30.4 & 40.8 & 24.7 & 24.4 & 29.8 \\ 
        40k & 33.9 & 36.9 & 25.5 & 27.3 & 19.5 & 26.7 & 39.7 & 28.3 & 40.7 & 25.4 & 22.6 & 29.7 \\ 
        160k & 34.7 & 37.7 & 26.2 & 28.2 & 19.9 & 27.0 & 40.9 & 31.3 & 40.7 & 25.7 & 24.9 & 30.7 \\ 
\midrule
\rowcolor{lightgray!30}     \multicolumn{13}{c}{\textbf{Bilingual Lower Layers [0-14] Fine-Tuning}} \\
\midrule
        10k & 33.4 & 36.2 & 25.6 & 27.1 & 18.4 & 24.2 & 32.8 & 23.1 & 42.1 & 25.5 & 18.5 & 27.9 \\ 
        20k & 33.1 & 36.9 & 25.4 & 27.2 & 18.3 & 24.1 & 33.1 & 25.6 & 41.8 & 25.1 & 19.3 & 28.2 \\ 
        40k & 33.9 & 37.8 & 25.6 & 27.5 & 19.2 & 25.7 & 34.8 & 25.8 & 41.2 & 25.3 & 21.7 & 29.0 \\ 
        160k & 35.5 & 39.3 & 27.0 & 30.1 & 19.7 & 25.9 & 39.4 & 31.3 & 39.9 & 25.2 & 24.6 & 30.7 \\ 
    \bottomrule
    \end{tabular}}
    \caption{The bilingual performance under different training strategies shows that fine-tuning the embedding layer performs as well as full fine-tuning in terms of bilingual performance. Interestingly, fine-tuning all lower layers does not yield additional gains. }
    \label{tab:training_all_low_layers}
\end{table*}

\section{More Analysis}
\label{appendix:analysis}

\noindent\textbf{The performance of Embed FT remains stable across reciprocal languages, regardless of the dataset being utilized.} As depicted in Table~\ref{tab:part-ft_fl}, the Embed FT strategy delivers performance that is competitive with the FT and LoRA strategies across all training sets: Alpaca-En, Alpaca-X, and Bilingual. Alpaca-En is a comprehensive English dataset with 52k instructions and demonstrations. Alpaca-X is derived from Alpaca-En through translation, with X denoting the target languages. The Bilingual dataset comprises 52k instruction data for translation tasks, based on the open-source Lego-MT dataset. Unlike the FT strategy, which updates all model parameters. Furthermore, it avoids the need for an additional model structure, like the LoRA strategy. This implies that Embed FT is a more effective strategy for activating multilingual capabilities.

In the Flores-101 dataset, the same evaluation metric, spBLEU, is used. Before calculating BLEU, all data is de-tokenized and sentence piece tokenization is applied to each language. This allows for a more accurate assessment of model quality on the long tail of low-resource languages. \textbf{NLU:} We evaluate various tasks to test different aspects of the model. These include XCOPA~\citep{ponti2020xcopa}, a multilingual common reasoning task supporting 11 languages; XStoryCloze~\citep{lin-etal-2022-shot}, a story completion task in 11 languages; XNLI~\citep{xnli}, a cross-lingual natural language inference task for 15 languages; PAWS-X~\citep{yang2019pawsx}, a paraphrase identification task in 7 languages; and MGSM~\citep{shi2022language}, a mathematical reasoning task in 11 languages.

\begin{table}[ht]
    \centering
    \footnotesize
    \resizebox{1\linewidth}{!}{
    \begin{tabular}{l|cccccc|c}
        \toprule
        \textbf{Models} & \textbf{XCOPA} & \textbf{MGSM} & \textbf{XStoryCloze} & \textbf{PAW-X} & \textbf{XNLI} & \textbf{Flores-101}  & \textbf{AVG.} \\
        \midrule
        Parrot-7B   & 54.2 & 3.7 & 56.1 & 56.5 & 39.0 & 25.2 & 46.9 \\
        LLaMA-7B  & 53.9 & 5.8 & 55.5 & 53.2 & 37.1  & 4.4 & 35.0  \\
        \midrule
        \multicolumn{8}{c}{LLaMA-7B + Alpaca-En} \\
        \midrule
        FT &  \textbf{54.5} & 4.5 & \textbf{57.6} & 57.1 & \textbf{40.3} & 28.2 & \textbf{48.4}\\
        LoRA  & 54.4 & 6.0 & 57.0 & 54.1 & 38.4 & 29.1 & 47.8\\
        Embed FT& 54.0 & \textbf{6.2} & 55.9 & 54.4 & 38.0 & 29.2 & 47.6 \\  
        \midrule
        \multicolumn{8}{c}{LLaMA-7B + Alpaca-X} \\
        \midrule
        FT & 54.4 & 4.9 & 57.2 & \textbf{57.1} & 40.2 & 28.0 & 48.4 \\
        LoRA & 54.5 & 5.6 & 57.0 & 53.8 & 38.3 & 28.0 & 47.4\\
        Embed FT  & 54.1 & 5.9 & 55.9 & 54.6 & 38.3 & 27.9 & 47.3 \\
        \midrule
        \multicolumn{8}{c}{LLaMA-7B + Bilingual} \\
        \midrule
        FT & 53.9 & 3.4 & 55.6 & 55.9 & 38.8 & 30.1 & 47.6\\
        LoRA & 54.3 & 4.7 & 55.9 & 54.3 & 38.0 & 31.1 & 47.6\\
        Embed FT & 54.3 & 4.7 & 55.9 & 54.3 & 38.0 & \textbf{31.4} & 47.7\\
        \bottomrule
    \end{tabular}
    }
    \caption{Comparative analysis of training strategies. XCOPA, MGSM, XStoryCloze, PAW-X and XNLI are natural language understanding tasks, evaluated on all languages with accuracy metric; Flores-101 is an NLG task, each score in the cell represents an average spBLEU, encompassing bilingual translation performances from en$\rightarrow$\{ro, es, de, ca, pt, da, no, bs\}.  The experimental result reveals that Embed FT can perform as well as another strategy.  }
\label{tab:part-ft_fl}
\end{table}

\noindent\textbf{Besides fine-tuning the embedding layer, adjusting the lower layers can also be effective.} To further investigate the functionality of the Embed FT strategy, we separately fine-tuned each layer of LLaMA using the Alpaca-En dataset and then tested on the Flores-101 en$\rightarrow$ro devtest. The layers of the LLaMA model, excluding the embedding layer, are numbered from 0 to 31, with 0 being the closest to the embedding layer and 31 being the furthest. The bilingual performance of en$\rightarrow$ro is illustrated in Table~\ref{tab:part-ft_fl}. Our experiments showed that fine-tuning the lower layers is just as effective as fine-tuning the embedding layer. However, we found that fine-tuning the higher layers did not produce satisfactory results.


\section{Used Scientific Artifacts}
Below lists scientific artifacts that are used in our work. For the sake of ethic, our use of these artifacts is consistent with their intended use.
\begin{itemize} [itemsep=1pt]
    \item \textit{Stanford Alpaca (Apache-2.0 license)}, a project that aims to build and share an instruction-following LLaMA model.
    \item \textit{Lego-MT (MIT license)}, a dataset for machine translation.
    \item \textit{Transformers (Apache-2.0 license)}, a framework that provides thousands of pretrained models to perform tasks on different modalities such as text, vision, and audio.
\end{itemize}

\section*{Acknowledge}
This work is partially supported by the National Key R\&D Program of China~(NO.2022ZD0160100).

\end{document}